\documentclass[10pt,twocolumn,letterpaper]{article}

\usepackage[pagenumbers]{cvpr} % To force page numbers, e.g. for an arXiv version

\usepackage{placeins}

\usepackage{graphicx}
\usepackage{amsmath}
\usepackage{amssymb}
\usepackage{booktabs}
\usepackage{multirow}
\usepackage{graphicx}  
\usepackage{booktabs}
\usepackage{multirow}
\usepackage{array}
\usepackage{afterpage}
\newcommand{\rot}[1]{\rotatebox[origin=c]{90}{#1}}

\definecolor{cvprblue}{rgb}{0.21,0.49,0.74}
\usepackage[pagebackref,breaklinks,colorlinks,allcolors=cvprblue]{hyperref}
\usepackage{color}
\def\paperID{18095} % *** Enter the Paper ID here
\def\confName{CVPR}
\def\confYear{2026}

\title{Basis-Oriented Low-rank Transfer for Few-Shot and Test-Time Adaptation}

\author{
Junghwan Park$^{1}$\quad
Woojin Cho$^{1}$\quad
Junhyuk Heo$^{1}$\quad
Darongsae Kwon$^{1}$\quad
Kookjin Lee$^{2}$\\[0.2cm]
$^{1}$TelePIX, \quad
$^{2}$Arizona State University\\[0.2cm]
{\tt\small \{junghwan, woojin, hjh1037, darong.kwon\}@telepix.net, kookjin.lee@asu.edu}
}

\begin{document}
\maketitle

%%%%%%%%% ABSTRACT
\begin{abstract}
Adapting large pre-trained models to unseen tasks under tight data and compute budgets remains challenging. Meta-learning approaches explicitly learn good initializations, but they require an additional meta-training phase over many tasks, incur high training cost, and can be unstable. At the same time, the number of task-specific pre-trained models continues to grow, yet the question of how to transfer them to new tasks with minimal additional training remains relatively underexplored. We propose BOLT (Basis-Oriented Low-rank Transfer), a framework that 
reuses existing fine-tuned models not by merging weights, but instead by extracting an orthogonal, task-informed spectral basis and adapting within that subspace.
In the offline phase, BOLT collects dominant singular directions from multiple task vectors and orthogonalizes them per layer to form reusable bases. 
In the online phase, we freeze these bases and train only a small set of diagonal coefficients per layer for the new task, yielding a rank-controlled update with very few trainable parameters. This design provides (i) a strong, training-free initialization for unseen tasks, obtained by pooling source-task coefficients—along with a lightweight rescaling step—while leveraging the shared orthogonal bases, and (ii) a parameter-efficient fine-tuning (PEFT) path that, in our experiments, achieves robust performance compared to common PEFT baselines as well as a representative meta-learned initialization. Our results show that constraining adaptation to a task-informed orthogonal subspace provides an effective alternative for unseen-task transfer.

\end{abstract}

%%%%%%%%% 1. INTRODUCTION
\section{Introduction}\label{sec:intro}
\FloatBarrier

\begin{figure}[t]
\centering
\includegraphics[width=0.75\columnwidth]{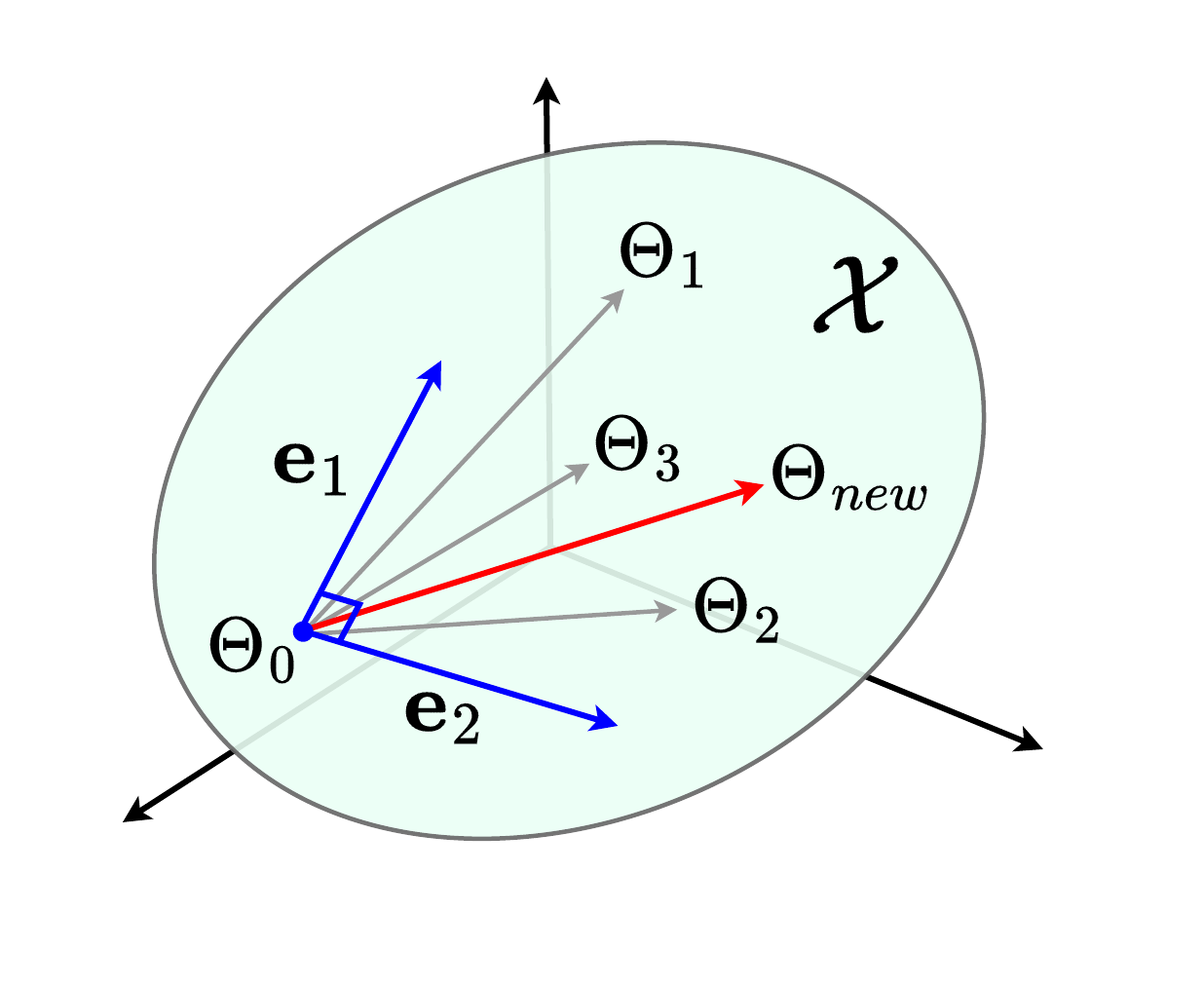}
\caption{\textbf{Conceptual view of BOLT}: Task-vectors $\{\Theta_i - \Theta_0\}_{i=1}^N$ can be represented in a common subspace $\mathcal{X}$ formed by orthogonal bases $\{\mathbf{e}_d\}_{d=1}^{r}$. By reusing the same subspace, the model can adapt to an unseen task vector $\{\Theta_\mathrm{new}-\Theta_0\}$ more quickly.}
\label{fig:concept_vis}
\end{figure}

Adapting large pre-trained models to new tasks and shifted data distributions is critical for real-world deployment, especially when facing unseen tasks with limited data~\cite{zhou2022learning, zhu2023not, liang2025comprehensive}. In such few-shot or test-time adaptation scenarios, the model’s initialization often plays a decisive role~\cite{chen2019closer, liang2025comprehensive} in whether learning will succeed.

Gradient-based meta-learning methods such as Model-Agnostic Meta-Learning (MAML)~\cite{finn2017model} were among the first to emphasize the importance of a good initialization by explicitly training models to rapidly adapt to novel tasks. However, these approaches involve expensive bi-level optimization and can be unstable to train~\cite{hospedales2021meta}, and require a dedicated meta-training phase over many tasks, which may be impractical in settings where new tasks continually emerge. This motivates developing methods that offer strong initialization for unseen tasks without requiring an additional meta-training stage, especially in settings where a library of fine-tuned models is already available.

At the same time, the ecosystem of pre-trained models is rapidly expanding, covering a diverse array of domains and tasks~\cite{goldblum2023battle, radford2021learning, dosovitskiy2020image}. This abundance presents an opportunity: instead of additional heavy training, one might transfer knowledge directly from existing models to new tasks. Prior work has explored merging or composing multiple fine-tuned models to combine their capabilities~\cite{matena2022merging, gargiulo2025task, jang2024model}. For example, a fine-tuned model can be represented as a task vector~\cite{ilharco2022editing}, and adding or averaging such vectors can blend learned behaviors. While these model-merging techniques can perform well on tasks that were part of their training set, they are not designed for entirely unseen tasks. A merged model essentially interpolates between known task solutions and tends to overfit to the training tasks, lacking a mechanism to generalize beyond them~\cite{wortsman2022model}. In fact, na\"ive task arithmetic often entangles disparate update directions, leading to interference and unpredictable behavior when tasks are misaligned. These limitations motivate a different approach to model transfer.
In this work, we pursue an alternative paradigm for adapting models to unseen tasks. If we have a library of models fine-tuned on diverse source tasks, the weight-space directions in which these models moved during fine-tuning may span a rich subspace for future tasks. Our insight is that these task-specific directions can be extracted and re-used as a basis for new tasks. Recent studies have shown that the dominant singular directions characterize the principal axes of the fine-tuning update, and that orthogonalizing such directions across tasks can mitigate interference. 

Building on this insight, we propose Basis-Oriented Low-Rank Transfer (BOLT), which adapts a pre-trained model to a new task without any explicit meta-training~\cite{hospedales2021meta} or model merging~\cite{yang2024model}. Instead of combining weights from different models, BOLT constructs an orthogonal spectral basis from existing fine-tuned models and uses this basis as the backbone for adapting to an unseen target task. A conceptual overview is shown in Figure~\ref{fig:concept_vis}.

\begin{figure}[t]
    \centering
    \includegraphics[width=0.8\columnwidth]{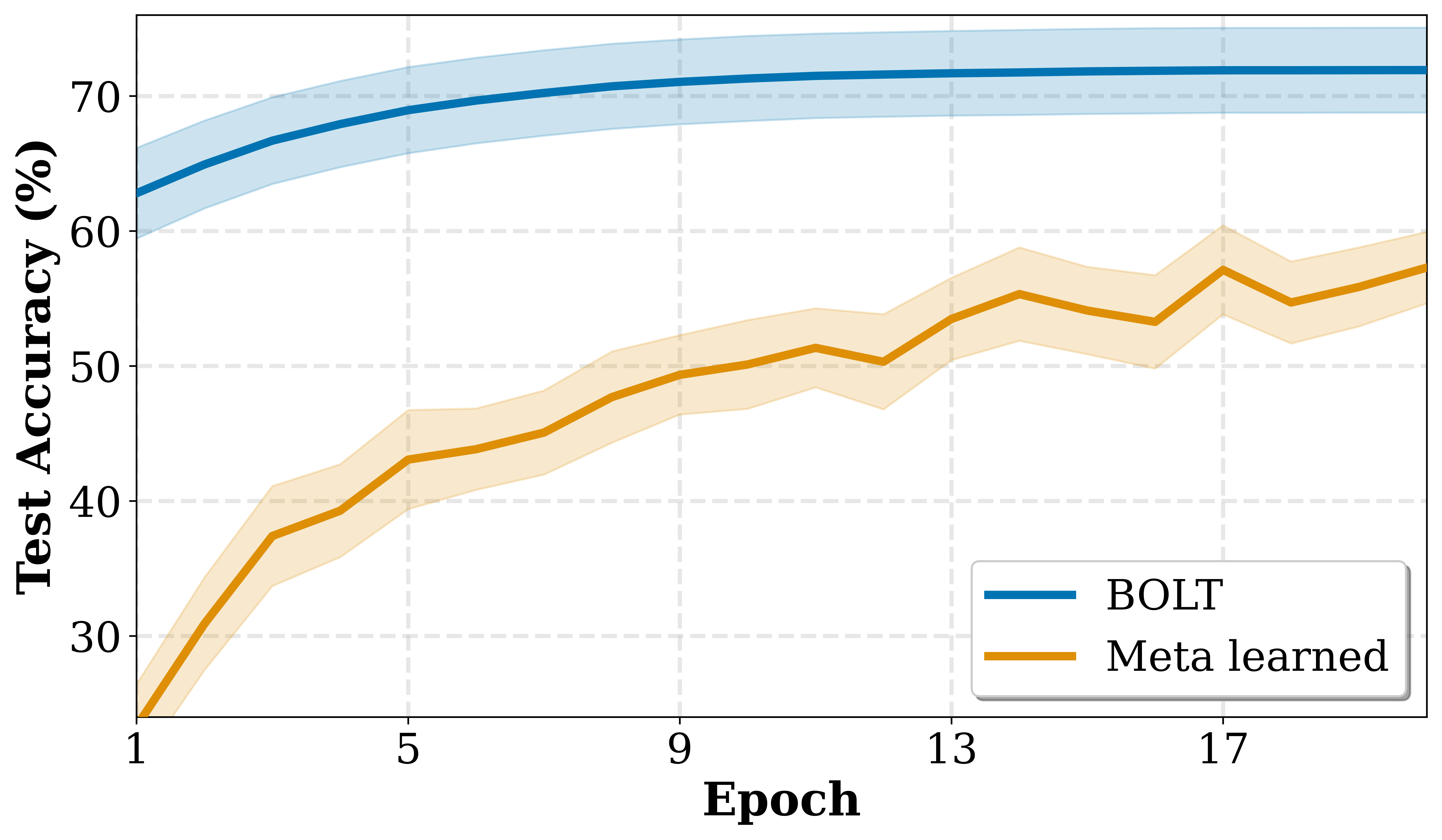}
    \caption{\textbf{Meta-learning vs.\ BOLT:} 
    few-shot adaptation curves comparing meta-learned initialization and BOLT. Our method reaches higher accuracy in fewer epochs, showing faster convergence than meta-learned initialization.}
    \label{fig:meta_vs_ours}
\end{figure}

Concretely, BOLT operates in two phases. In an offline phase, we collect the task vectors from multiple source models and perform singular value decomposition (SVD) to extract their major directions.
We then orthogonalize these directions to form a shared spectral basis for each layer. In the subsequent online phase, the new task is learned by freezing this orthogonal basis and training only a small set of coefficients. 
The resulting weight update for the unseen task is constrained to lie in the span of those basis directions – equivalently, the update is diagonal in the shared spectral basis.
This design keeps the number of trainable parameters extremely low and confines the update to a subspace spanned by prior tasks, providing explicit control over the update’s effective rank and reducing the risk of overfitting to spurious patterns.
We validate BOLT on challenging few-shot image classification benchmarks and label-free test-time adaptation scenarios. Without any additional meta-training, BOLT achieves robust performance than conventional fine-tuning or parameter-efficient tuning baselines. 
As an additional comparison, Figure~\ref{fig:meta_vs_ours} reports few-shot adaptation results on 17 unseen tasks using a ViT-B/32 backbone, where BOLT’s initialization outperforms meta-learned initialization. In summary, our contributions are:
\begin{itemize}
    \item \textbf{Orthogonal multi-task bases}: We build task-informed, layer-wise orthogonal bases by collecting the top singular directions from multiple fine-tuned models and orthogonalizing them. This provides a stable shared coordinate system for later adaptation.

    \item \textbf{Strong initialization without training}: By projecting task vectors into the shared spectral basis and pooling their coefficients—with a lightweight rescaling step—we obtain a training-free initialization that already performs well on unseen tasks.

    \item \textbf{Generalization on few-shot and test-time adaptation}: Comprehensive evaluation on few-shot, out-of-domain (OOD), and label-free test-time adaptation across general and remote-sensing datasets, showing that BOLT consistently matches or surpasses strong PEFT with far fewer task-specific parameters.
\end{itemize}

%%%%%%%%% 2. RELATED WORK
\section{Related Work}

\paragraph{Parameter-efficient adaptation.}
When data and compute are limited, a central challenge is enabling adaptation with minimal changes to the backbone. Recent work addresses this by constraining where learnable parameters are introduced. Low-rank modules such as LoRA~\cite{hu2022lora} and its decomposed variants~\cite{liu2024dora} inject compact update factors into linear layers, while modulation- and adapter-based methods~\cite{lian2022scaling,chen2022adaptformer} adjust activations or insert small bottlenecks to provide controlled capacity. Prompt-based tuning~\cite{jia2022visual} instead steers the backbone using a few learnable tokens. In the CLIP~\cite{radford2021learning} setting, feature-space approaches—including linear probes, TIP-Adapter~\cite{zhang2022tip}, and LP++~\cite{huang2024lp++}—avoid weight updates entirely. Despite differing in where capacity is added, these methods all restrict adaptation to a lightweight and well-structured subspace.

\paragraph{Task vectors and composition.}
A complementary line of work edits models directly in weight space by representing fine-tuned checkpoints as task vectors, enabling arithmetic operations such as averaging, interpolation, and tangent-space editing~\cite{ilharco2022editing,ortiz2023task}. While this reveals that independently fine-tuned models often lie in shared basins, na\"ive combinations can entangle incompatible update directions, prompting structure-aware approaches that use curvature-based weighting~\cite{matena2022merging} or conflict-aware sparsification~\cite{yadav2023ties}. More recent analyses decompose task vectors into layer-wise singular directions~\cite{gargiulo2025task}, providing a means to diagnose interference and merge updates along decorrelated axes, while anisotropic scaling methods~\cite{zhang2024knowledge} learn how to emphasize or suppress specific components. Collectively, these methods expose the compositional structure of fine-tuning updates, though they primarily reweight existing directions rather than building new coordinate systems for unseen tasks.

\paragraph{Subspace methods and orthogonalization.}
A related line of work controls where learning occurs by shaping the subspace of allowable updates. In continual learning, projecting gradients away from previously spanned directions mitigates forgetting~\cite{zeng2019continual,saha2021gradient}, and later studies analyze how the choice and frequency of such projections influence robustness~\cite{zhao2023rethinking,yang2023data}. Orthogonalization also arises in representation analysis, where similarity metrics, whitening, and decorrelated normalization help disentangle feature directions and reduce redundancy~\cite{raghu2017svcca,kornblith2019similarity,huang2018decorrelated}. In parameter space, permutation or orthogonal alignment reconciles symmetries before model averaging, improving merge fidelity~\cite{ainsworth2022git}. Overall, these approaches demonstrate that enforcing orthogonality—on gradients, representations, or weights—can reduce interference and motivate learning within a shared, decorrelated subspace.

\paragraph{Initialization for Unseen Tasks.}
A central determinant of performance on unseen tasks is the starting point from which adaptation proceeds. While gradient-based meta-learning popularized learning such initializations, it typically requires a separate meta-training phase over many tasks~\cite{finn2017model, nichol2018first, cho2023hypernetwork}. A complementary line of work emphasizes structure in the update space, using low-rank or orthogonalized coordinates to stabilize early learning. Building on this perspective, we propose a training-free initializer and show that pooling and lightly rescaling the spectral coefficients of source-task updates provides a compact starting point that performs well on unseen tasks. Conceptually, our approach is related to meta-learning in that both aim to construct useful initializations for new tasks, but BOLT obtains such initial states analytically from a collection of fine-tuned models instead of introducing a separate meta-optimization stage.

%%%%%%%%% 3. PRELIMINARIES
\section{Preliminaries}\label{sec:preliminary}
Before introducing our framework, we formalize the problem setting, and clarify how task-specific updates are represented and combined.
\subsection{Setting and Task Vectors}
We consider a pre-trained model with weights $\Theta_0$, and a set of $N$ related source tasks for which fine-tuned models $\{\Theta_i\}_{i=1}^N$ are available. Each $\Theta_i$ is obtained by fine-tuning $\Theta_0$ on task $i$. We define the task vector for task $i$ as the difference between the fine-tuned and original weights:
\begin{equation}
\Delta_i \;=\; \Theta_i - \Theta_0 .\label{eq:task}
\end{equation}
This $\Delta_i$ represents the full update that adapts the pre-trained model to task $i$. 
For each layer $\ell$, suppose the weight parameters in $\Theta_0$ form a matrix of dimensions $m_\ell \times n_\ell$ and let $M_i^{(\ell)} = \Theta_i^{(\ell)} - \Theta_0^{(\ell)}$ be the corresponding layer update for task $\ell$. This layer-wise view aligns with prior analyses that use singular vector to characterize task-specific changes and allows us to directly apply standard linear algebra tools.

% For each layer, suppose the weight parameters form a matrix of dimensions $m_\ell \times n_\ell$ in $\Theta_0$. Then the corresponding layer update for task $i$ can be viewed as a matrix $M_i^{(\ell)}$ (with $M_i^{(\ell)} = \Theta_i^{(\ell)} - \Theta_0^{(\ell)}$). Focusing on matrix-shaped layers allows us to leverage linear algebra tools; any non-matrix parameters (e.g. biases or normalization parameters) can be handled separately. This layer-wise view preserves the structure of each weight matrix and aligns with prior analyses that use singular vectors to characterize task-specific changes.
% We adopt a layer-wise perspective by reshaping weight parameters into matrices and handling non-matrix terms separately. This approach preserves the weight-matrix structure and enables the use of standard matrix methods.

\subsection{Task Arithmetic and Its Limitations}
A simple way to combine knowledge from multiple tasks is to merge their weight updates. We first define the merged model $\Theta_{agg}$ as follows:
\begin{equation}
\Theta_{{agg}} \;=\; \Theta_0 + \sum_{i=1}^{N} \alpha_i \Delta_i.\label{eq:agg}
\end{equation}
Here, $\Delta_i$ is the task vector of task $i$, and $\alpha_i$ are blending coefficients. This task arithmetic directly composes the fine-tuning updates from different tasks. 
While easy to compute, na\"ive weight merging as in Eq.~\eqref{eq:agg} can suffer from interference between tasks. 
If the dominant update directions of tasks are not mutually aligned, the $\sum_i \alpha_i \Delta_i$ will entangle disparate directions and degrade performance. In practice, merged models often exhibit unpredictable behavior unless the tasks are very compatible. More importantly, merging alone is not sufficient for entirely new target tasks. The combination in Eq.~\eqref{eq:agg} can only blend the behaviors of existing task updates ${\Delta_i}$, and when the target task departs from these sources, there is no principled way to choose coefficients ${\alpha_i}$ that produce the required behavior. In particular, once a merge is chosen, there is no explicit adaptation step for the new task. In contrast, our approach treats these directions as a shared basis and learns task-specific coefficients from the target data rather than relying on a single static merge.

\subsection{Singular Directions and Low-Rank Structure}\label{sec:low_rank}
Although a task update $\Delta_i$ has the same dimensionality as the original parameters $\Theta_0$, in practice the change it induces is much more structured. Empirically, most layers do not need to move in all directions of the weight space at once; instead, the update tends to concentrate its energy in a few dominant directions. To make this precise, for a given layer $\ell$, we look at the matrix-shaped update $M_i^{(\ell)}$ corresponding to task $i$ at that layer and apply a thin SVD:
\begin{equation}
M_i^{(\ell)} \;=\; U_i^{(\ell)} \Sigma_i^{(\ell)} V_i^{(\ell)\top},\label{eq:svd}
\end{equation}
where $U_i^{(\ell)}$ and $V_i^{(\ell)}$ collect the left and right singular directions of the update, and $\Sigma_i^{(\ell)}$ contains the singular values sorted in decreasing order. When only a few singular values in $\Sigma_i^{(\ell)}$ are large, Eq.~\eqref{eq:svd} shows that $M_i^{(\ell)}$ is well-approximated by a low-dimensional subspace spanned by those dominant singular directions. This is the key property we later exploit when we aggregate singular directions from multiple source tasks and orthogonalize them into the shared bases.

\begin{figure*}[t]
\centering
\includegraphics[width=2.0\columnwidth]{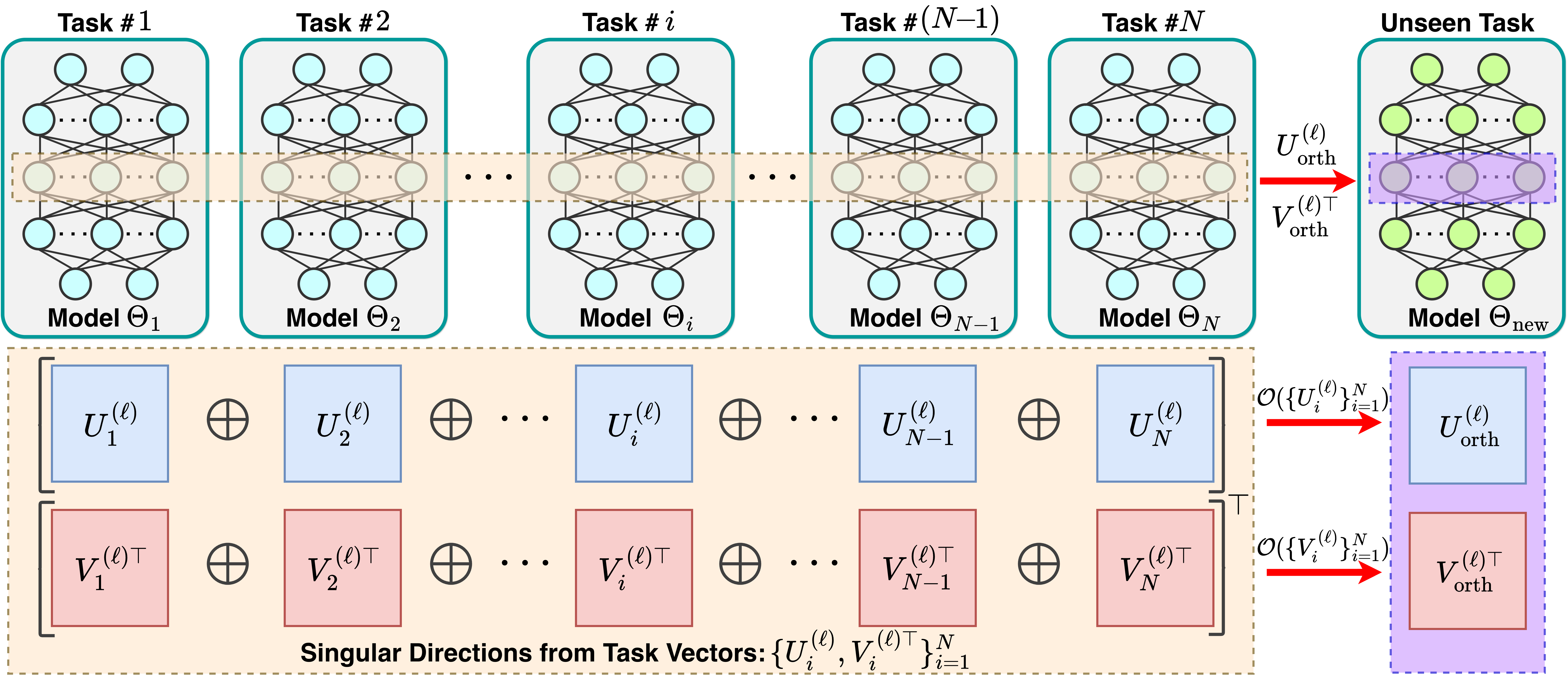}
\caption{\textbf{Overall BOLT pipeline}: for task vectors $\{\Theta_i\}_{i=1}^N$, we extract layer-wise SVDs and orthogonalize them to obtain shared bases $\{U_{\mathrm{orth}}, V_{\mathrm{orth}}\}$. These fixed bases are later reused to construct weights for an unseen task with only small diagonal parameters.}\label{fig:archi}
\end{figure*}

\subsection{Orthogonalization and Basis Construction}
\label{sec:orth-basis}
As shown in Section~\ref{sec:low_rank}, task-specific layer updates $M_i^{(\ell)}$ often concentrate their energy in a few singular directions (Eq.~\eqref{eq:svd}). This suggests that, across many source tasks, we can collect those dominant directions and use them to span a common subspace for future adaptation.

For each layer $\ell$, suppose we have $N$ source tasks and their matrix-shaped updates $\{M_i^{(\ell)}\}_{i=1}^N$. We first compute a thin SVD for each and keep only the top $k_i$ singular directions. We then stack these directions across tasks to make
\begin{equation}
\begin{aligned}
U_{\mathrm{stack}}^{(\ell)} \!&=\!
\begin{bmatrix}
U_1^{(\ell)}(:,1\!:\!k_1) & \!\!\!\! \cdots \!\!\!\!& U_{N}^{(\ell)}(:,1\!:\!k_N)
\end{bmatrix}\in \mathbb{R}^{m_\ell \times r},\\
V_{\mathrm{stack}}^{(\ell)} \!&=\!
\begin{bmatrix}
V_1^{(\ell)}(:,1\!:\!k_1) & \!\!\!\! \cdots \!\!\!\! & V_{N}^{(\ell)}(:,1\!:\!k_N)
\end{bmatrix}\in \mathbb{R}^{n_\ell \times r}.
\end{aligned}\label{eq:stack}
\end{equation}
So that $U_{\mathrm{stack}}^{(\ell)} \in \mathbb{R}^{m_\ell \times r}$ and $V_{\mathrm{stack}}^{(\ell)} \in \mathbb{R}^{n_\ell \times r}$, where $r = \sum_{i=1}^N k_i$ is the total number of collected directions.

Because these stacked directions may still be correlated across tasks, we orthogonalize them to obtain the final task-informed, layer-wise bases $U_{\mathrm{orth}}^{(\ell)}$ and $V_{\mathrm{orth}}^{(\ell)}$, defined as
\begin{equation}
U_{\mathrm{orth}}^{(\ell)} \;=\; \mathcal{O}\!\big(U_{\mathrm{stack}}^{(\ell)}\big),
\qquad
V_{\mathrm{orth}}^{(\ell)} \;=\; \mathcal{O}\!\big(V_{\mathrm{stack}}^{(\ell)\top}\big)^{\top},
\end{equation}
where $\mathcal{O}(\cdot)$ denotes the SVD-based orthogonalization operator. 
These matrices serve as the shared spectral coordinates in which all subsequent target-task updates are represented.
Concretely, we apply a single whitening step
\begin{equation}
\begin{split}
U_{\mathrm{orth}}^{(\ell)} &= U_{\mathrm{stack}}^{(\ell)}
\big( {U_{\mathrm{stack}}^{(\ell)}}^\top U_{\mathrm{stack}}^{(\ell)} + \varepsilon I \big)^{-\tfrac{1}{2}},
\qquad \varepsilon > 0, \\
V_{\mathrm{orth}}^{(\ell)} &= V_{\mathrm{stack}}^{(\ell)}
\big( {V_{\mathrm{stack}}^{(\ell)}}^\top V_{\mathrm{stack}}^{(\ell)} + \varepsilon I \big)^{-\tfrac{1}{2}}, \qquad \varepsilon > 0.
\end{split}
\label{eq:orth}
\end{equation}
with $\varepsilon>0$ for numerical stability. Following prior work~\cite{gargiulo2025task}, we compute $U_{\mathrm{stack}}^{\ell}$ and $V_{\mathrm{stack}}^{\ell}$ via the thin SVD $U_{\mathrm{stack}}^{\ell}=\Psi_{u}\Sigma_u \Phi_u^\top$ and $V_{\mathrm{stack}}^{\ell}=\Psi_{v}\Sigma_v \Phi_v^\top$. We then set $U_{\mathrm{orth}}=\Psi_u \Phi_u^\top$ and $V_{\mathrm{orth}}=\Psi_v \Phi_v^\top$. 

All subsequent adaptation in Section~\ref{sec:method} will keep $U_{\mathrm{orth}}^{(\ell)}$ and $V_{\mathrm{orth}}^{(\ell)}$ fixed and will learn only how much to scale each basis direction.

% Given a stack of directions $X$ from multiple tasks, we denote by
% \begin{equation}
% \mathcal{O}(X) \;\in\; \mathbb{R}^{d\times r}
% \quad\text{with}\quad
% \mathcal{O}(X)^\top \mathcal{O}(X) = I_r
% \end{equation}
% an orthogonalization of its columns. In practice, $\mathcal{O}$ can be realized by a numerically stable map such as a QR-based or polar-based normalization of $X$. Working in $\mathcal{O}(X)$ reduces directional overlap before any learning on the target task.

% \paragraph{Orthogonalization.}
% Given a stack $X\in\mathbb{R}^{d\times r}$ (full column rank), we obtain a column-orthonormal basis by the polar decomposition,
% \begin{equation}
% Q \;=\; X\,(X^\top X + \varepsilon I)^{-1/2},\quad \varepsilon>0,
% \end{equation}
% and use $Q$ as the shared basis. In practice we compute $Q$ via the thin SVD $X=U\Sigma V^\top$ and set $Q=UV^\top$ (the orthogonal Procrustes factor).

% We follow a two-phase strategy: an offline phase builds per-layer orthogonal bases from source tasks, and an online phase adapts to a new task by learning a small set of coefficients in the shared coordinates. This separation mirrors successful designs that decouple representation construction from lightweight adaptation.

\section{Proposed Method}\label{sec:method}
In this section, we now describe how BOLT performs adaptation using the shared orthogonal bases. Figure~\ref{fig:archi} summarizes the overall pipeline, and the following subsections describe each step mathematically. 

\subsection{Projection to The Shared Coordinates}
Let $M^{(\ell)}_i$ denote the parameter update of source task $i$ at layer $\ell$, defined as in Sections~\ref{sec:intro}–\ref{sec:preliminary}. For layer $\ell$, we omit the index for clarity and denote $M := M^{(\ell)}_i$, $U := U^{(\ell)}_{\mathrm{orth}}$, $V := V^{(\ell)}_{\mathrm{orth}}$.
Because $U$ and $V^\top$ are column- and row-orthonormal, respectively, any update $M$ can be projected onto the
task-informed subspace by
\begin{equation}
    S \;=\; U^\top M V, \quad S \in \mathbb{R}^{r \times r},
    \label{eq:proj}
\end{equation}
which measures how strongly $M$ activates each pair of shared directions obtained in Eqs.~\eqref{eq:stack}-\eqref{eq:orth}.
Note that this step does not re-define the bases; it only re-expresses $M$ in the already-fixed coordinates.

\subsection{Diagonal Reconstruction Problem}
Our goal in Section~\ref{sec:method} is not to keep the full $r \times r$ matrix in Eq.~\eqref{eq:proj}, but to find the best
diagonal matrix inside this subspace that reconstructs $M$ when mapped back to the original space.
We therefore consider the least-squares problem
\begin{equation}
    \min_{D} \; \| M - U D V^\top \|_F^2,
    \label{eq:diag-ls}
\end{equation}
where $D \in \mathbb{R}^{r \times r}$ is constrained to be diagonal, i.e., $D_{ij}=0$ for $i \neq j$. Eq.~\eqref{eq:diag-ls} asks for the diagonal coefficients in the shared basis that best match the offline update $M$.
To make Eq.~\eqref{eq:diag-ls} explicit, insert and subtract $USV^\top$ inside the Frobenius norm:
\begin{align}
 &\| (M - USV^\top)+ U(S - D)V^\top \|_F^2\label{eq:split-1} \\
=  &\| M - USV^\top \|_F^2 + \| U(S - D)V^\top \|_F^2,\label{eq:split-2}
\end{align}
The cross term in Eq~\eqref{eq:split-1} vanishes because the two components lie in orthogonal subspaces:
\begin{equation}
    U^\top (M - USV^\top) V \;=\; S - S \;=\; 0.
\end{equation}
The first term in Eq.~\eqref{eq:split-2} is independent of $D$, so minimizing Eq.~\eqref{eq:diag-ls} is equivalent to
\begin{equation}
    \min_{D} \; \| U(S - D)V^\top \|_F^2.
    \label{eq:reduced}
\end{equation}
At this point we use the orthonormality already established in Section~\ref{sec:preliminary}. When a matrix already lives in the span of $U$ and $V$, the Frobenius norm is preserved. Thus, the term in  Eq.~\eqref{eq:reduced} can be expressed as follows:
\begin{equation}
    \| U(S - D)V^\top \|_F^2 \;=\; \| S - D \|_F^2.
    \label{eq:norm-preserve}
\end{equation}
Therefore, the diagonal extraction problem further reduces to the following form:
\begin{equation}
    % \min_{D \;\text{diagonal}} \; \| S - D \|_F^2.
    \min_{D} \; \| S - D \|_F^2.
    \label{eq:final-diag}
\end{equation}

\subsection{Closed-form Diagonal and Coefficient Vector}
Expanding the Frobenius norm in Eq.~\eqref{eq:final-diag} yields Eq.~\eqref{eq:frob_sep}:
\begin{equation}
    \| S - D \|_F^2
    = \sum_{j \neq k} S_{jk}^2 + \sum_{j} (S_{jj} - D_{jj})^2.\label{eq:frob_sep}
\end{equation}
The off-diagonal part $\sum_{j \neq k} S_{jk}^2$ does not depend on $D$, so the minimizer is obtained by matching every
diagonal entry. In other words, the optimal diagonal simply copies the diagonal of $S$:
\begin{equation}
    D_{jj}^\star = S_{jj} \quad \forall j.\label{eq:closed_diag}
\end{equation}
Consequently, we can write
\begin{equation}
    D^\star = \operatorname{diag}(\mathbf{s}), \qquad
    \mathbf{s} := \operatorname{diag}(S) \in \mathbb{R}^r.
    \label{eq:coeff-vector}
\end{equation}
The $r$-dimensional vector $\mathbf{s}$ is the only information from $M$ that we will carry over into the online phase.

\subsection{Pooling Across Multiple Source Tasks}
\begin{figure}[t]
\centering
\includegraphics[width=1.0\columnwidth]{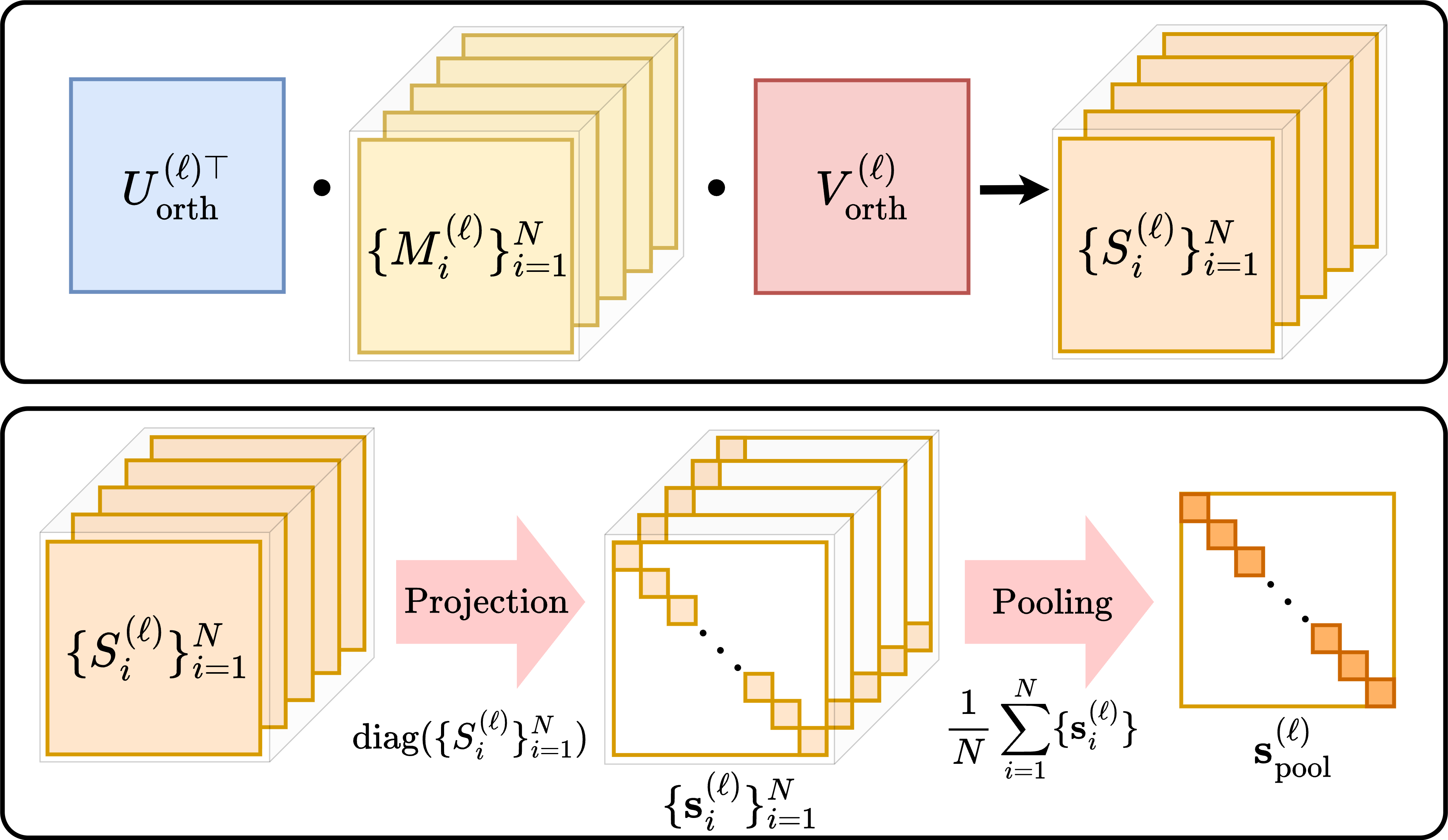}
\caption{\textbf{Initialization of the diagonal coefficients}: each source task is projected onto $\{U_{\mathrm{orth}}, V_{\mathrm{orth}}\}$ to obtain its best diagonal $\{\mathbf{s}_i^{\ell}\}_{i=1}^N$, and these per-task diagonals are pooled to form $\mathbf{s}_{\mathrm{pool}}^{\ell}$ for a new task. This provides a data-free, task-informed starting point.}
\label{fig:init_archi}
\end{figure}

When we have $N$ tasks, we simply repeat the projection Eq.~\eqref{eq:proj} and the diagonal extraction
Eq.~\eqref{eq:coeff-vector} for each task:
\begin{equation}
    S_i = U^\top M_i V,\!\! \quad \!\!\mathbf{s}_i = \operatorname{diag}(S_i) \in \mathbb{R}^r,
    \!\! \quad \!\! i = 1, \dots, N,
    \label{eq:multi-source}
\end{equation}
These $\{\mathbf{s}_i\}_{i=1}^N$ are all expressed in the same orthonormal coordinates
because $U$ and $V$ were fixed offline (i.e. Eqs.~\eqref{eq:stack}-\eqref{eq:orth}).
A simple, data-free initializer for a new/unseen task is then the component-wise mean
\begin{equation}
    \mathbf{s}_{\mathrm{pool}} \;=\; \frac{1}{N} \sum_{i=1}^N \mathbf{s}_i,
    \label{eq:mean-init}
\end{equation}
Figure~\ref{fig:init_archi} summarizes pooling, where each source update is projected onto the shared bases to form $S_i$, and then the resulting diagonals $\{\mathbf{s}_i\}_{i=1}^N$ are averaged to obtain $\mathbf{s}_{\mathrm{pool}}$ for the online phase.

\paragraph{Scalar rescaling of the diagonal.}
After the data-free mean initializer in Eq.~\eqref{eq:mean-init}, we apply a single global scaling factor to all diagonal coefficients, selecting it via a brief sweep on the training set. For each candidate $\alpha\in\mathcal{A}$ (a small finite set, e.g., $\{1,3,5,7,10\}$), we form
\begin{equation}
\tilde{\mathbf{s}}^{(\ell)}(\alpha)
= \alpha\,\mathbf{s}_{\mathrm{pool}}^{(\ell)}
\quad \text{for all layers } \ell ,
\label{eq:s_init}
\end{equation}
compose the merged parameters 
\begin{equation}
\Theta_0+\sum_{\ell}U_{\mathrm{orth}}^{(\ell)}\operatorname{diag}(\tilde{\mathbf{s}}^{(\ell)}(\alpha))V_{\mathrm{orth}}^{(\ell)\top},
\end{equation}
and evaluate accuracy on a few mini-batches of the training data. We then select
\begin{equation}
\hat{\alpha}\;=\;\underset{{\alpha\in\mathcal{A}}}{\arg\max}\;\mathrm{Acc}\big(\Theta_0,\{\tilde{\mathbf{s}}^{(\ell)}(\alpha)\};\mathcal{D}_{\mathrm{train}}\big),
\end{equation}
and set the initialization to $\mathbf{s}_{0}^{(\ell)}=\hat{\alpha}\,\mathbf{s}_{\mathrm{pool}}^{(\ell)}$ for all $\ell$. This step does not modify the orthogonal bases and adds negligible cost, yet often stabilizes early adaptation by matching the overall scale of the diagonal update to the target task.

\subsection{Online Adaptation in The Shared Subspace}
At adaptation time, we keep the task-informed bases from Section~\ref{sec:preliminary} fixed and learn only the spectral coefficients. For a new task, the admissible update at layer $\ell$ is
\begin{equation}
\Delta_{\mathrm{new}}^{(\ell)}(\mathbf{s}^{(\ell)})=
U_{\mathrm{orth}}^{(\ell)}\operatorname{diag}(\mathbf{s}^{(\ell)})V_{\mathrm{orth}}^{(\ell)\top}.
\end{equation}
where $\mathbf{s}^{(\ell)}$ is the only learnable parameter at that layer. This parameterization restricts the update to the spectral diagonal of the shared basis, so that $\operatorname{rank}(\Delta_\mathrm{new}^{(\ell)}(\mathbf{s}^{(\ell)})) \le r$ 
% \begin{equation}
    % \operatorname{rank}\!\big(\Delta^{(\ell)}(s^{(\ell)})\big) \;\le\; r,
% \end{equation}
and forces the target-task update to remain inside the task-informed subspace spanned by $\{U^{(\ell)}_{\mathrm{orth}}, V^{(\ell)}_{\mathrm{orth}}\}$.
The adapted model for the new task can then be written as
\begin{equation}
\Theta(\mathbf{s}) = \Theta_0 + \sum_{\ell} \Delta_{\mathrm{new}}^{(\ell)}(\mathbf{s}^{(\ell)}),
\end{equation}
where all $\mathbf{s}^{(\ell)}$ are initialized from the pooled and rescaled coefficients in Eq.~\eqref{eq:s_init}.
In this way, BOLT performs online adaptation by optimizing only a few number of spectral coefficients per layer, dramatically reducing the number of task-specific parameters while still allowing effective adaptation in the shared orthogonal subspace.

% \subsection{Objectives and regularization}
% Few-shot regimes minimize a supervised loss on the small labeled set; label-free test-time regimes minimize an unsupervised objective such as entropy under strong augmentations. A quadratic spectral penalty stabilizes extreme scalings:
% \begin{equation}
% \min_{\{\mathbf{s}^{(\ell)}\}}\;\; \mathcal{L}\big(\theta(\mathbf{s})\big) \;+\; \lambda \sum_{\ell}\big\|\mathbf{s}^{(\ell)}\big\|_2^2 .
% \end{equation}

% \subsection{Rank control and safeguards}
% The spectral width $r$ is capped by layer dimensions and chosen via $k_i$. If a layer’s numerical rank is zero, its update is set to zero. Non-matrix parameters use an averaged initialization and remain frozen. When a higher-capacity update is desired, $r$ can be increased by raising $k_i$ or by orthonormal completion of the basis up to $\min(m_\ell,n_\ell)$, trading parameter efficiency for flexibility.
%%%%%%%%% 5. EXPERIMENTS

\begin{table*}[t]
\centering
\small
\setlength{\tabcolsep}{8pt}
\caption{Few-shot accuracy (\%) averaged over three CLIP backbones (ViT-B/16, ViT-B/32, ViT-L/14) and all datasets within each domain. 
Left: general-domain; right: remote-sensing. 
Best and second-best per column are in \textbf{bold} and \underline{underlined}.}
\label{tab:fewshot_general_rs_multicol}
\begin{tabular}{lccccccccccc}
\specialrule{1pt}{2pt}{2pt}
& \multicolumn{5}{c}{General datasets} & \multicolumn{5}{c}{Remote-sensing datasets} \\
\cmidrule(lr){2-6}\cmidrule(lr){7-11}
Method
       & 1-shot & 2-shot & 4-shot & 8-shot & 16-shot
       & 1-shot & 2-shot & 4-shot & 8-shot & 16-shot \\
\midrule
Zero-shot CLIP
& \multicolumn{5}{c}{\phantom{0}60.74}
& \multicolumn{5}{c}{\phantom{0}58.59} \\
\cmidrule(lr){1-11}
Linear Probe 
& 61.08 & 61.09 & 61.11 & 61.16 & 61.23 
& 58.87 & 58.95 & 59.19 & 59.67 & 60.70 \\
LoRA
& 60.97 & 61.40 & 64.86 & 65.34 & 69.57
& \underline{70.12} & \underline{73.91} & \underline{81.14} & \underline{86.31} & \underline{90.76} \\
TIP
& 61.49 & 61.74 & 62.44 & 63.05 & 63.78
& 60.36 & 62.01 & 64.94 & 68.49 & 73.26 \\
LP++
& 43.34 & 54.11 & 60.88 & 67.83 & 72.09
& 51.85 & 67.59 & 77.36 & 84.00 & 88.47 \\
aTLAS
& \underline{66.34} & \underline{67.37} & \underline{71.31} & \underline{73.41} & \underline{74.44}
& 69.87 & 73.68 & 77.86 & 81.38 & 86.24 \\
BOLT (ours)
& \textbf{71.30} & \textbf{72.67} & \textbf{74.90} & \textbf{76.72} & \textbf{78.46}
& \textbf{80.77} & \textbf{83.86} & \textbf{87.59} & \textbf{89.92} & \textbf{91.86} \\
\specialrule{1pt}{2pt}{2pt}
\end{tabular}
\end{table*}

\begin{figure*}[t]

% ---- 공통 레전드 ----
\centering
\begin{minipage}{0.45\linewidth}
  \centering
  \includegraphics[width=\linewidth]{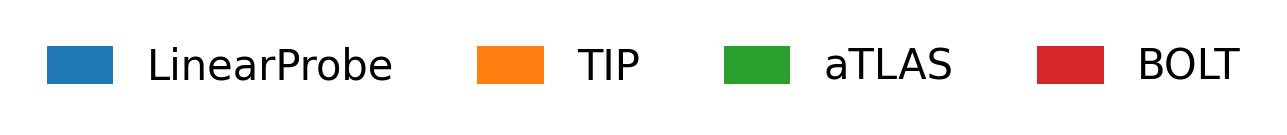}
\end{minipage}

\vspace{0.8em}

\begin{minipage}{0.24\linewidth}
  \centering
  \includegraphics[width=\linewidth]{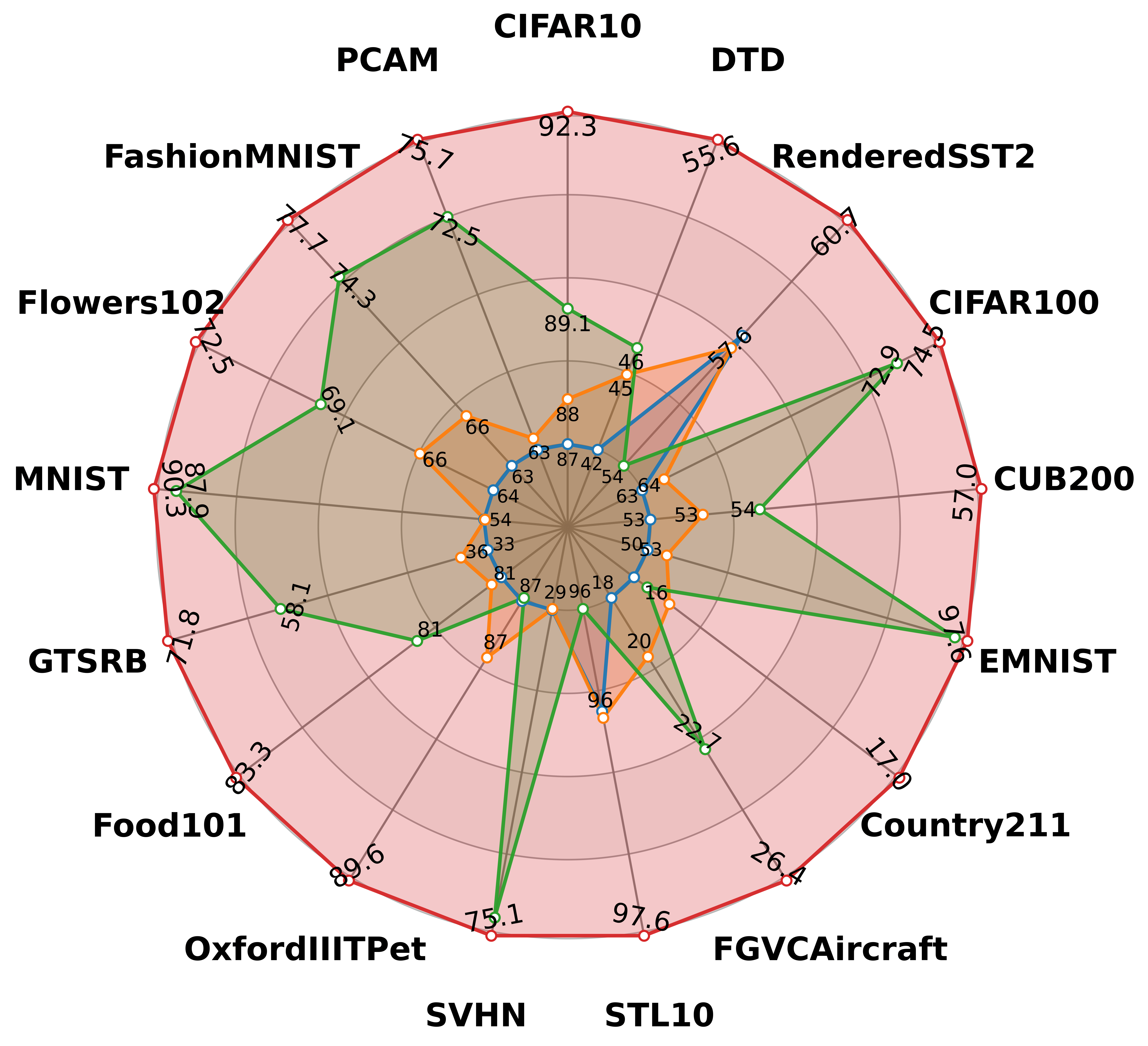}
  (a) General, 4-shot
\end{minipage}%  <-- 줄바꿈 방지
\hfill
\begin{minipage}{0.24\linewidth}
  \centering
  \includegraphics[width=\linewidth]{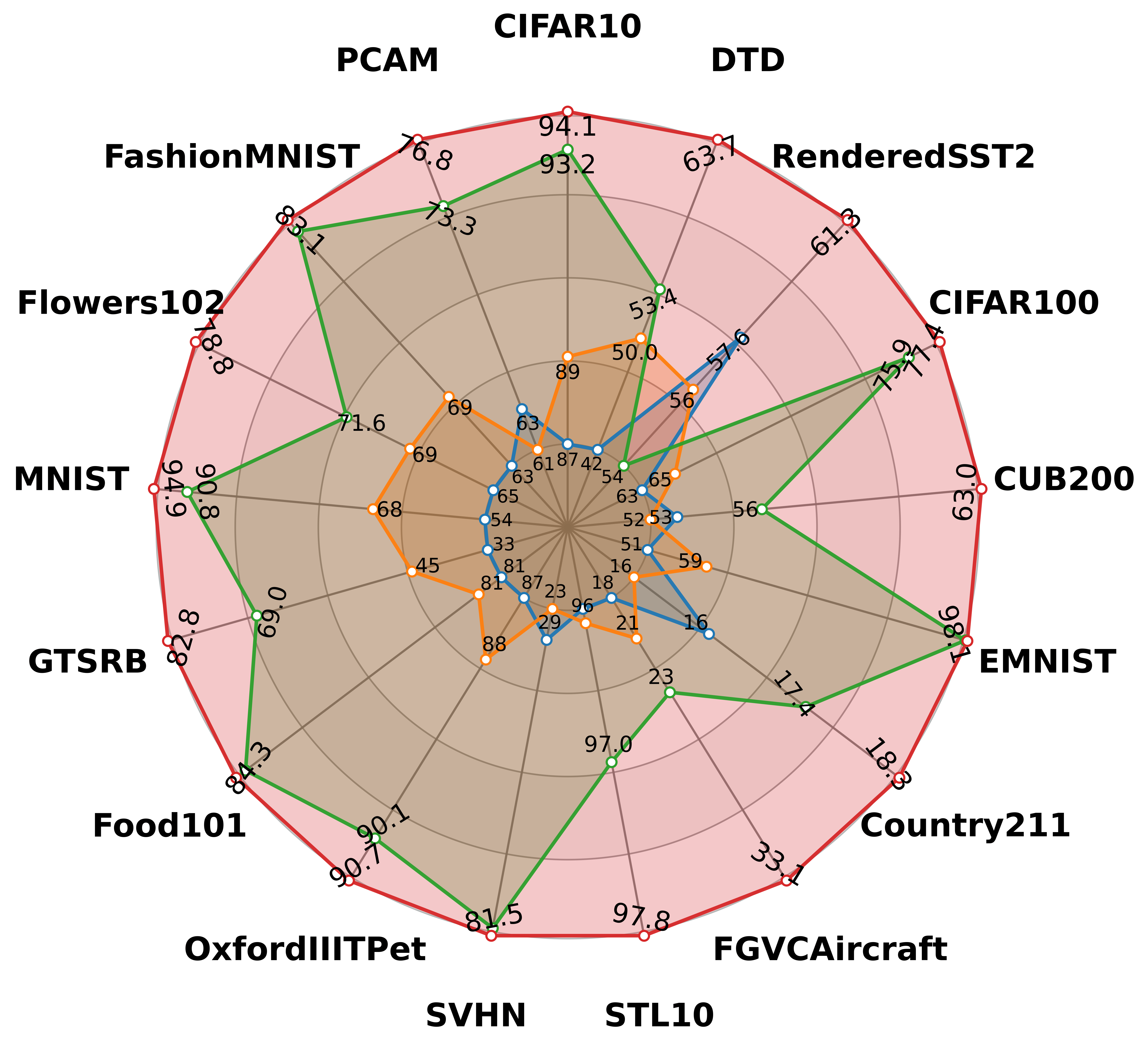}
  (b) General, 16-shot
\end{minipage}%
\hfill
\begin{minipage}{0.24\linewidth}
  \centering
  \includegraphics[width=\linewidth]{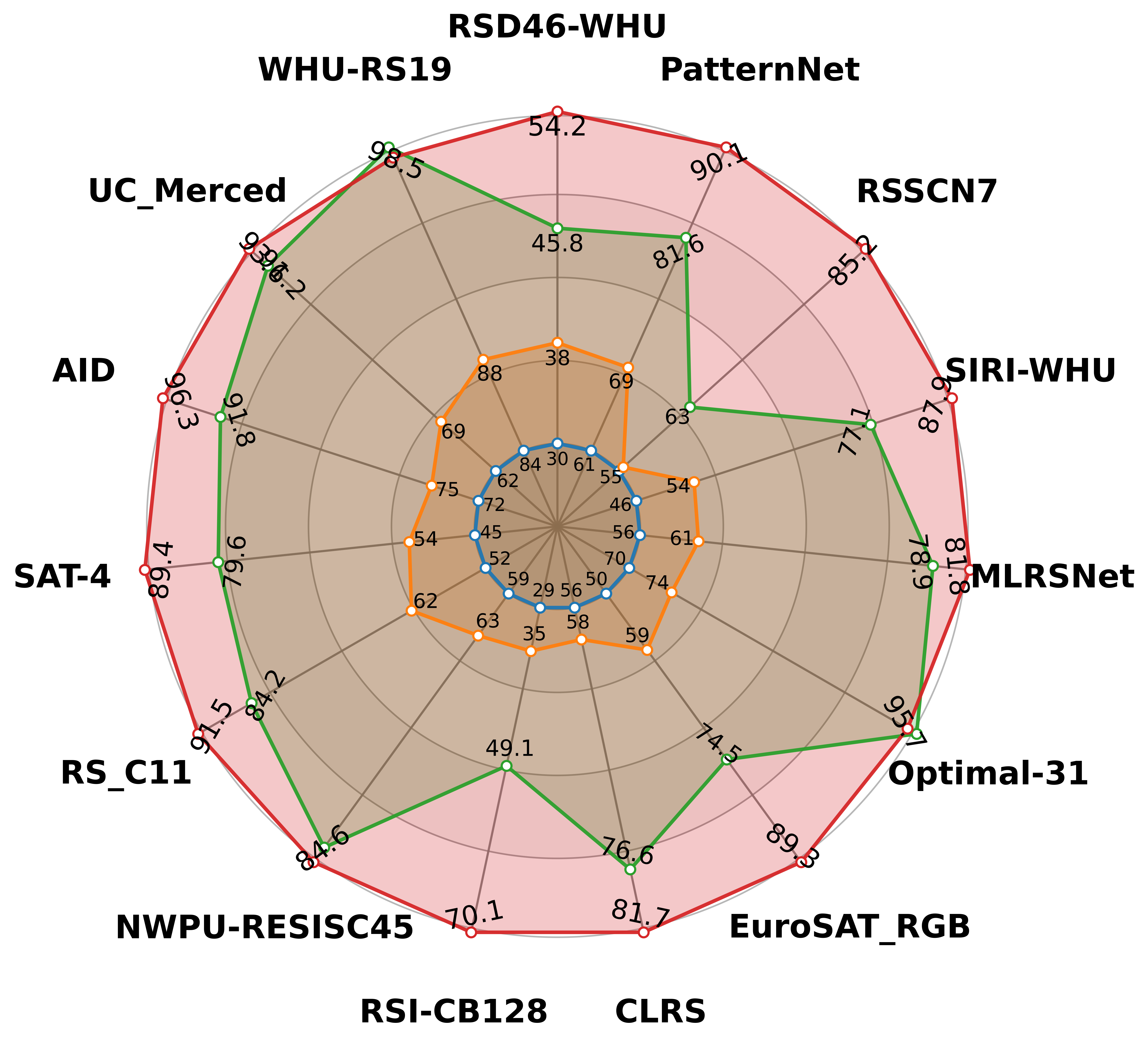}
  (c) Remote sensing, 4-shot
\end{minipage}%
\hfill
\begin{minipage}{0.24\linewidth}
  \centering
  \includegraphics[width=\linewidth]{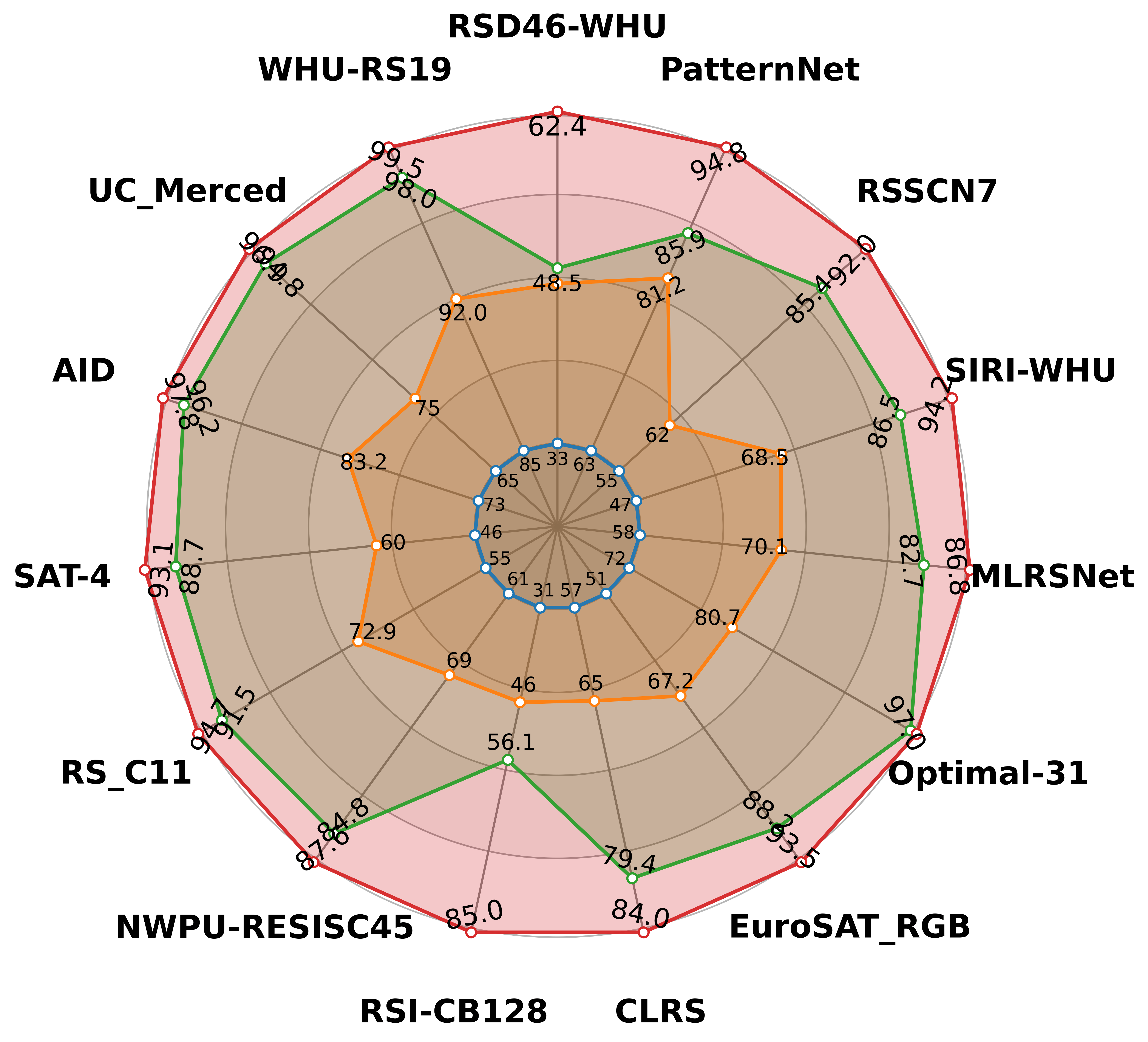}
  (d) Remote sensing, 16-shot
\end{minipage}

\caption{Few-shot accuracy at 4 and 16 shots using the ViT-B/32 backbone across general and remote-sensing datasets.}
\label{fig:four_panel_general_rs}
\end{figure*}

\section{Experiments}
We evaluate BOLT across few-shot, OOD, and test-time adaptation settings to validate its generality and efficiency. This section outlines the benchmarks and implementation details, followed by quantitative comparisons and ablations.
\subsection{Benchmarks and setup}
\paragraph{Tracks.}
We evaluate three complementary tracks: (i) few-shot classification with $k\!\in\!\{1,2,4,8,16\}$ labeled samples per class, (ii) OOD robustness on ImageNet-V2/A/R/Sketch, and (iii) label-free test-time adaptation (TTA) under distribution shift.

\paragraph{Datasets.}
We evaluate our method on two disjoint families of tasks. 
\textbf{(i) General-domain tasks} include DTD~\cite{cimpoi2014describing}, GTSRB~\cite{stallkamp2011german}, MNIST~\cite{lecun1998mnist}, SVHN~\cite{netzer2011reading}, STL10~\cite{coates2011analysis}, OxfordIIITPet~\cite{parkhi2012cats}, Flowers102~\cite{nilsback2008automated}, CIFAR100~\cite{krizhevsky2009learning}, PCAM~\cite{veeling2018rotation}, CIFAR10~\cite{krizhevsky2009learning}, Food101~\cite{bossard2014food}, FashionMNIST~\cite{xiao2017fashion}, RenderedSST2~\cite{socher2013recursive}, EMNIST~\cite{cohen2017emnist}, FGVCAircraft~\cite{maji2013fine}, CUB200~\cite{wah2011caltech}, and Country211~\cite{radford2021learning}. 
\textbf{(ii) Remote-sensing tasks} include AID~\cite{xia2017aid}, CLRS~\cite{li2020clrs}, EuroSAT\_RGB~\cite{helber2019eurosat}, MLRSNet~\cite{qi2020mlrsnet}, NWPU-RESISC45~\cite{cheng2017remote}, Optimal-31~\cite{wang2018scene}, PatternNet~\cite{zhou2018patternnet}, RS\_C11~\cite{zhao2016feature}, RSD46-WHU~\cite{xiao2017high}, RSI-CB128~\cite{li2020rsi}, RSSCN7~\cite{zou2015deep}, SAT-4~\cite{zhu2016bag}, SIRI-WHU~\cite{zhu2016bag}, UC\_Merced~\cite{yang2010bag}, and WHU-RS19~\cite{xia2010structural}. 

We report results separately for these two families because they differ markedly in their visual statistics and transfer behavior. General-domain datasets contain natural or object-centric imagery with relatively homogeneous texture and illumination, whereas remote-sensing datasets consist of aerial scenes characterized by greater spatial complexity, scale variation, and distribution shifts induced by environmental or atmospheric factors. Evaluating them jointly would obscure domain-specific effects, while treating them separately allows us to examine how adaptation behaves under distinct visual regimes. Notably, remote-sensing imagery forms a particularly challenging stress test for few-shot and low-rank adaptation, since its domain gap from pre-trained CLIP representations is substantially larger than in conventional vision benchmarks.

\paragraph{Baselines.}
We compare BOLT against a range of adaptation methods. Zero-shot CLIP~\cite{radford2021learning} classifies images directly from text prompts without any task-specific training, while a linear probe~\cite{radford2021learning} fits a linear classifier on frozen CLIP features and serves as a lightweight adaptation baseline. Tip-Adapter~\cite{zhang2022tip} performs training-free few-shot adaptation using a key--value cache built from support examples, and LP++~\cite{huang2024lp++} strengthens linear probing through improved feature normalization and optimization. Among parameter-efficient tuning methods, LoRA~\cite{hu2022lora} injects low-rank updates into weight matrices, and aTLAS~\cite{zhang2024knowledge} learns anisotropic scaling coefficients over task vectors to improve compositional transfer.

\paragraph{Training and evaluation.}
We benchmark BOLT on three CLIP backbones—ViT-B/32, ViT-B/16, and ViT-L/14—across few-shot and test-time adaptation settings to assess robustness under diverse data regimes. 
In all experiments, BOLT performs layer-wise retention of up to 12 singular directions, capturing dominant update structure while maintaining a highly compact model (about 8k learnable parameters). The spectral bases are constructed by aggregating all source-task vectors within each domain.
Implementation details, hyperparameters, and extended results are provided in the supplementary material.

\subsection{Few-Shot Adaptation}
Few-shot learning aims to adapt a pre-trained model to a new task using only a handful of labeled samples—$k$ images per class for a $k$-shot setting. 
Following standard practice, we adapt a frozen CLIP encoder to each target dataset using only $k\!\in\!\{1,2,4,8,16\}$ class-balanced examples while keeping the spectral basis fixed.
As summarized in Table~\ref{tab:fewshot_general_rs_multicol}, BOLT achieves the best performance across both the general and remote-sensing domains, showing clear advantages even under extremely limited supervision.
In few-shot settings, the improvement margin is particularly pronounced, highlighting the effectiveness of spectral basis adaptation for data-scarce scenarios.
While aTLAS learns anisotropic scaling coefficients on existing task vectors, reweighting known directions within a linear subspace, our method instead constructs a compact diagonal spectral basis derived from multi-task singular directions. 
By decorrelating these directions into an orthogonal basis, BOLT mitigates interference during adaptation and achieves stable, data-efficient performance, as shown in Table~\ref{tab:fewshot_general_rs_multicol}.
In addition, aTLAS requires all task vectors to be stored and optimized jointly, which increases memory usage, whereas BOLT builds a shared orthogonal basis once and reuses it across tasks, achieving much higher memory efficiency without sacrificing performance.

% In the remote-sensing subset, improvements are even more pronounced. 
% Unlike general-domain datasets, remote-sensing imagery exhibits high intra-class variance and domain shifts due to atmosphere, sensor, and scale differences. 
% The spectral diagonalization in BOLT naturally regularizes adaptation under such shifts, as each basis direction captures a disentangled mode of task variation. 
% This compact representation enables robust generalization even when only a few labeled samples are available.

Figure~\ref{fig:four_panel_general_rs} illustrates per-dataset results for the ViT-B/32 backbone at 4- and 16-shot highlighting dataset-specific performance. 
BOLT consistently ranks highest across diverse datasets in both general and remote-sensing domains, confirming its strong adaptability across heterogeneous distributions. 
% Notably, at 4-shot, BOLT already recovers most of its 16-shot performance, indicating that the precomputed spectral bases provide highly informative priors for few-shot learning.

\begin{table}[t]
\centering
\small
\setlength{\tabcolsep}{4pt}
\caption{OOD accuracy (\%) using ViT-B/32 backbone at 16-shot. 
Results are reported on ImageNet OOD variants.}
\label{tab:ood_vitb32_k16}
\resizebox{\columnwidth}{!}{
\begin{tabular}{l c c c c}
\specialrule{1pt}{2pt}{2pt}
Method       & ImageNet-A & ImageNet-R & ImageNet-S & ImageNet-V2 \\
\midrule
Zero-shot CLIP & 14.76 & 50.95 & 38.92 & 52.91  \\
Linear Probe &\underline{15.13} & 51.49 & 39.48 & 54.72 \\
LoRA         & 10.28 & 42.16 & 36.76 & 54.51 \\
TIP          & 14.64 & 50.56 & 38.62 & 52.00 \\
LP++         & 14.66 & 51.92 & 39.01 & 54.22 \\
aTLAS        & 14.87 & \underline{52.21} & \underline{39.83} & \underline{55.29} \\
BOLT (ours)  & \textbf{15.88} & \textbf{53.85} & \textbf{41.26} & \textbf{55.69} \\
\specialrule{1pt}{2pt}{2pt}
\end{tabular}
}
\end{table}

\subsection{OOD Robustness}
We evaluate robustness under distribution shift using the ViT-B/32 backbone with 16-shot supervision. All methods are trained on ImageNet~\cite{russakovsky2015imagenet} and then evaluated under distribution shift on ImageNet-A~\cite{deng2009imagenet}, ImageNet-R~\cite{hendrycks2021many}, ImageNet-S (Sketch)~\cite{wang2019learning}, and ImageNet-V2~\cite{recht2019imagenet}. Table~\ref{tab:ood_vitb32_k16} shows that BOLT delivers the strongest overall performance, achieving the highest accuracy on all four OOD datasets.

While linear probing remains a competitive baseline, lightweight adaptation methods such as LP++ and TIP exhibit notable degradation under these challenging shifts. LoRA performs particularly poorly, likely due to its large number of learnable parameters, which makes it more susceptible to overfitting in low-data OOD scenarios. In contrast, BOLT maintains stable accuracy across all benchmarks, suggesting that its layer-wise spectral parameterization—derived from task-vector subspaces—transfers effectively even when the input distribution changes substantially. These results indicate that BOLT provides strong OOD generalization without introducing heavy task-specific modules or increasing model capacity.

\begin{table}
  \caption{TTA results on general-domain datasets, reporting accuracy (\%) averaged over all datasets, where the best and second-best methods are highlighted in \textbf{bold} and \underline{underlined}, respectively.}
  \label{tab:tta_general}
  \centering
  {\small
  \begin{tabular}{@{}lccc@{}}
    \specialrule{1pt}{2pt}{2pt}
    Method & ViT-B/16 & ViT-B/32 & ViT-L/14 \\
    \midrule
    Zeroshot CLIP  & 61.55 & 56.87 & 67.90 \\
    Layer Norm     & \underline{67.63} & \underline{62.68} & \underline{74.08} \\
    aTLAS          & 61.56 & 56.87 & 67.92 \\
    BOLT (ours)    & \textbf{69.92} & \textbf{67.74} & \textbf{74.11} \\
    \specialrule{1pt}{2pt}{2pt}
  \end{tabular}
  }
\end{table}

\subsection{Test-Time Adaptation (TTA)}
We study test-time adaptation~\cite{liang2025comprehensive} in a fully label-free setting, where the model receives only unlabeled images from the target dataset and adapts offline with access to the full split.
Our approach uses a unified variant of unsupervised FixMatch (UFM)~\cite{sohn2020fixmatch} tailored to BOLT. For each input we generate weak and strong augmentations, use the weak view for sharpened pseudo-labels, and mark high-confidence predictions as trusted with fixed targets. The remaining samples are treated as unlabeled and optimized via a consistency loss between strong-view predictions and pseudo-labels with confidence-based masking, over a few epochs under a cosine learning-rate schedule.

Table~\ref{tab:tta_general} reports the resulting TTA performance on general-domain datasets for three CLIP backbones.
Across all backbones, BOLT consistently improves over the zero-shot CLIP model and existing baselines, with the largest gains on the smaller ViT-B/32 backbone, suggesting that spectral diagonal adaptation helps compensate limited capacity by exploiting unlabeled target data more effectively.
Overall, coupling BOLT’s spectral basis with a UFM-style objective yields strong label-free adaptation under distribution shift while keeping the encoder architecture fixed and updating only low-dimensional spectral coefficients instead of introducing new task-specific modules.

\subsection{Ablation Study}

\begin{figure}[t]
  \centering
  % --- (a) Ablation on rank r ---
  \begin{subfigure}[t]{0.485\columnwidth}
    \centering
    \includegraphics[width=\linewidth]{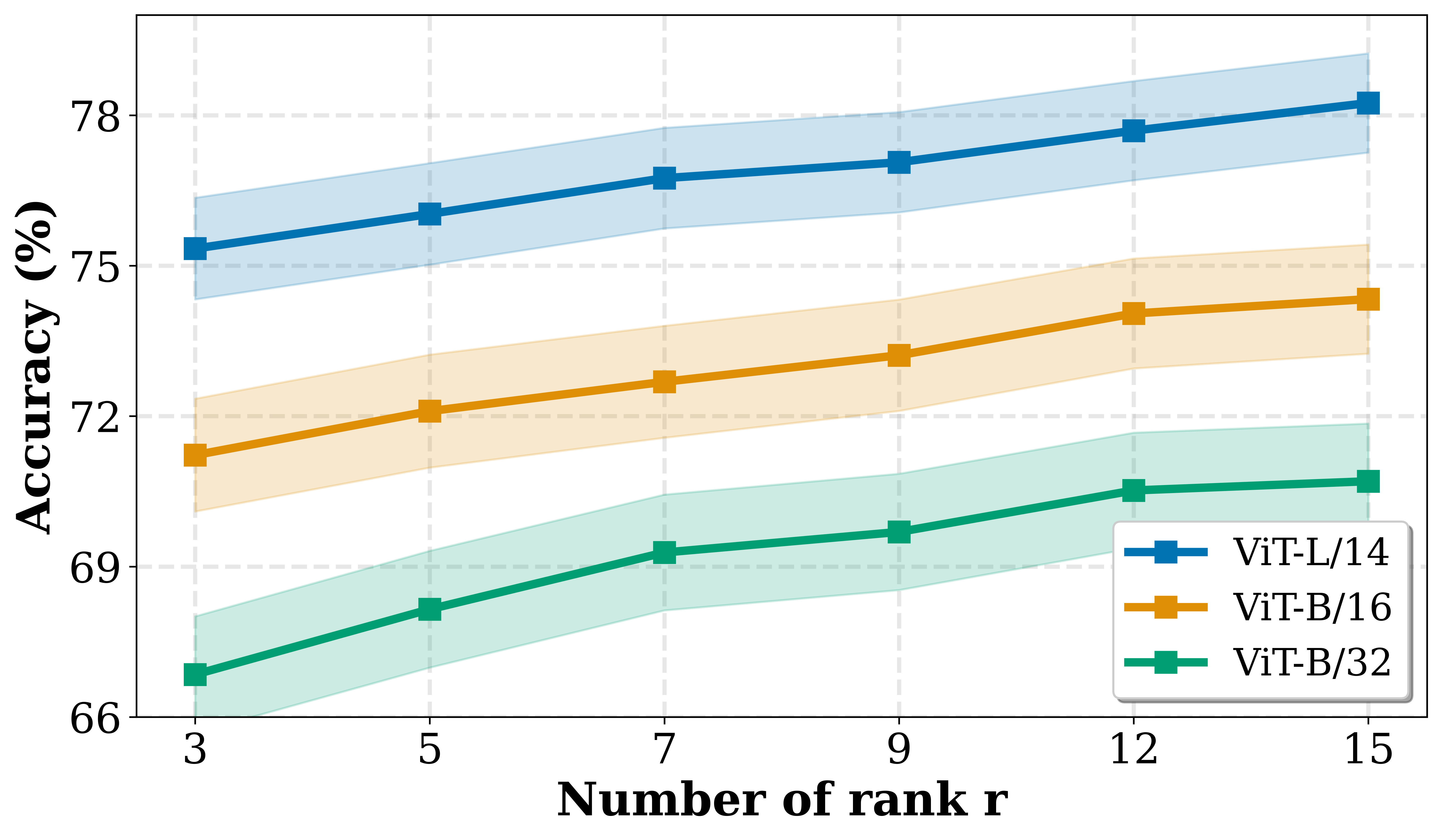}
    \caption{Effect of \# rank $r$.}
    \label{fig:rank_ablation}
  \end{subfigure}
  \hfill
  % --- (b) Number of source tasks ---
  \begin{subfigure}[t]{0.485\columnwidth}
    \centering
    \includegraphics[width=\linewidth]{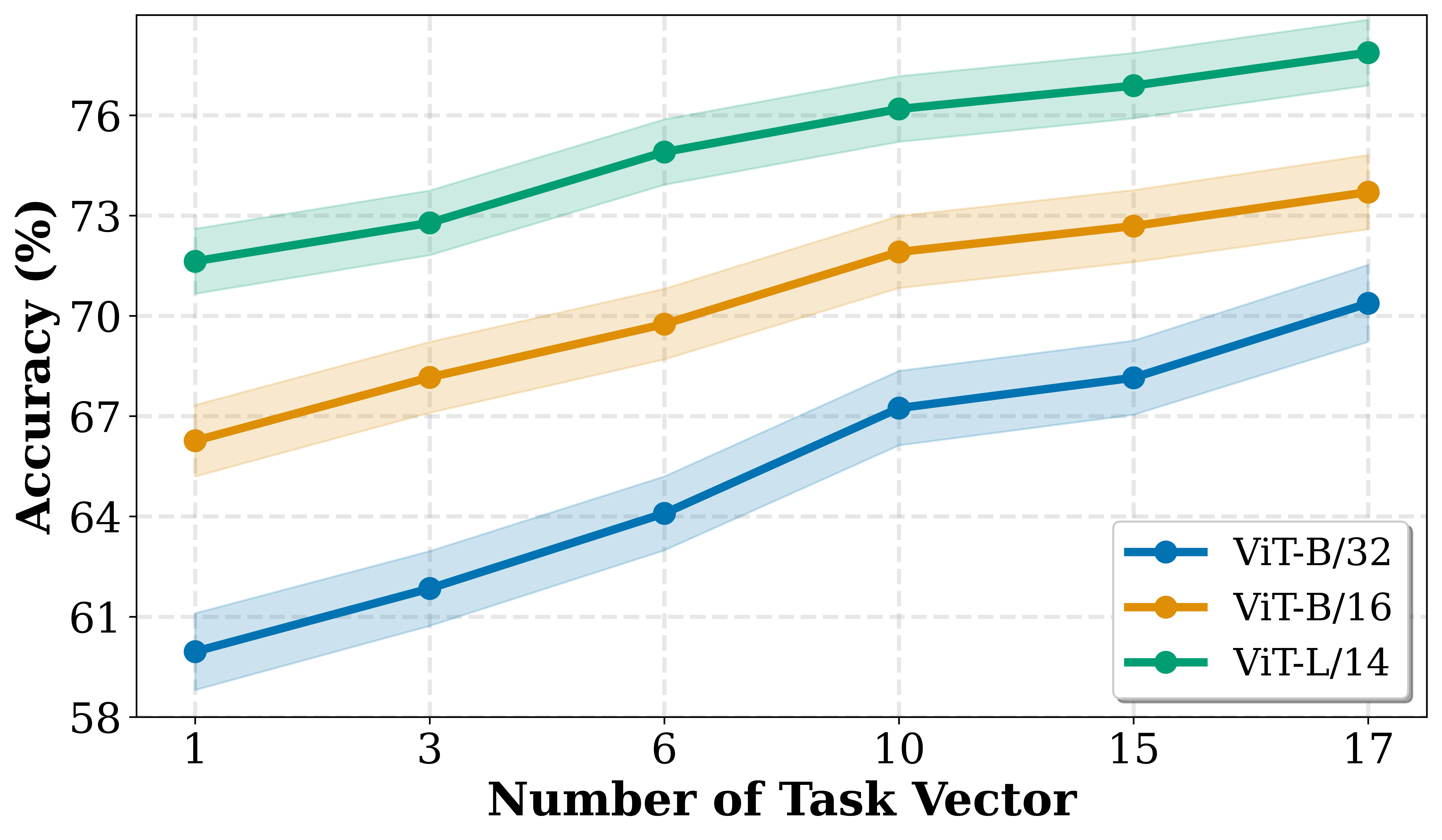}
    \caption{Effect of \# task vectors.}
    \label{fig:numtasks_plot}
  \end{subfigure}
  \caption{
  \textbf{Ablation studies on spectral basis construction and adaptation.}
  (a) Accuracy vs.\ \# rank $r$, showing that performance saturates after a small subspace size. 
  (b) Accuracy vs.\ \# task vectors used to build the shared basis, indicating that a compact yet diverse set of sources suffices.}
  \label{fig:ablation_combined}
\end{figure}

\label{sec:ablation}
We study two factors in constructing the spectral basis: the layer-wise rank $r$ and the number of source task vectors. All ablations use the 16-shot setting, average over three seeds, and train only the diagonal spectral coefficients; source vectors are selected at random.

\paragraph{Effect of the rank $r$.}
Using all source task vectors, we vary the number of singular vectors kept per layer. As shown in Figure.~\ref{fig:rank_ablation}, all backbones benefit from increasing $r$, with the smaller ViT-B/32 showing the strongest sensitivity at low ranks. Larger models (ViT-B/16, ViT-L/14) remain stable even with small $r$, and performance generally saturates around $r\approx12$.

\paragraph{Effect of the number of source task vectors.}
Fixing $r=12$, we vary the number of task vectors used to form the basis. Figure.~\ref{fig:numtasks_plot} shows rapid gains when adding the first few sources—again most pronounced for ViT-B/32—after which performance plateaus as the subspace becomes sufficiently well spanned.

%%%%%%%%% 7. CONCLUSION
\section{Conclusion}
We presented BOLT, a spectral adaptation framework that constructs layer-wise orthogonal bases from task vectors and adapts new tasks by learning only diagonal coefficients in this shared subspace. This design removes the reliance on any meta-training stage and instead provides a compact, controllable update space grounded in task-informed spectral coordinates. Empirically, BOLT delivers strong few-shot performance, robust out-of-domain generalization, and effective label-free test-time adaptation across multiple CLIP backbones, consistently matching or surpassing strong parameter-efficient tuning baselines. These results demonstrate that orthogonal, task-informed spectral coordinates offer a simple and scalable mechanism for low-parameter transfer in large pre-trained models.

\clearpage 

{
    \small
    \bibliographystyle{ieeenat_fullname}
    \bibliography{main}
}

\clearpage
\appendix

%%%%%%%%%%%%%%%%%%%%%%%%%%%%%%%%%%%%%%%%%%%%%%%%%%%%%%%%%%%%
\section{Implementation Details}
%%%%%%%%%%%%%%%%%%%%%%%%%%%%%%%%%%%%%%%%%%%%%%%%%%%%%%%%%%%%

This section provides additional information about
(i) the compute environment used in all experiments, and
(ii) training and hyperparameter settings for few-shot, OOD, and test-time adaptation,
(iii) high-level pseudo-code for the BOLT pipeline.

%-----------------------------------------------------------
\subsection{Compute Environment}
%-----------------------------------------------------------

All experiments were conducted on the same compute environment:
\begin{itemize}
  \item CPU: INTEL(R) XEON(R) PLATINUM 8570
  \item GPU: NVIDIA B200
  \item OS: Ubuntu 22.04.5 LTS
  \item Python: 3.11.12
  \item PyTorch: 2.7.0
\end{itemize}
Few-shot, OOD, and TTA experiments all use this software stack.

%-----------------------------------------------------------
\subsection{Backbones and Trainable Parameters}
%-----------------------------------------------------------

We use CLIP vision encoders ViT-B/32, ViT-B/16, and ViT-L/14 as frozen backbones.
The CLIP text encoder is kept fixed and is only used to construct zero-shot classifiers.
For BOLT, the only trainable parameters are the layer-wise spectral diagonal coefficients
in the shared orthogonal basis; all CLIP weights and bases remain frozen.
For baselines (Linear Probe, LoRA, TIP, LP++, aTLAS) we follow their standard
parameterizations while keeping the backbone frozen and sharing the same data splits.

%-----------------------------------------------------------
\subsection{Initialization and Alpha Grid Search}
%-----------------------------------------------------------

Before any sigma-only adaptation (few-shot, OOD, or TTA), we perform a common
initialization step:

\begin{itemize}
  \item \textbf{Basis construction:} from a set of fine-tuned models on source tasks,
        we derive task vectors and compute SVD-based orthogonal bases
        $U_\text{orth}, V_\text{orth}$ for each weight matrix, together with an initial
        diagonal coefficient vector $\sigma$ per module.
  \item \textbf{Sigma parametrization:} each matrix update is represented as
        $\Delta(\sigma) = U_\text{orth} \,\mathrm{diag}(\sigma)\, V_\text{orth}$,
        and the collection of all $\sigma$ forms the only trainable parameters.
  \item \textbf{Global scaling:} we run a short grid search over a global scale
        $\alpha \in \{1, 3, 5, 7, 10\}$, evaluate the merged encoder
        $\Theta(\alpha) = \Theta_0 + \Delta(\alpha \cdot \sigma)$ on the 
        train data loader, and select the best $\hat{\alpha}$, which is then fixed for the rest of training.
\end{itemize}

This procedure is shared across all adaptation scenarios and provides a strong,
data-informed initialization for the sigma parameters.

%-----------------------------------------------------------
\subsection{Few-shot Adaptation Settings}
%-----------------------------------------------------------

In the few-shot setting, each dataset in turn is treated as a held-out target task,
while the remaining datasets are used to construct the spectral basis. For each
target dataset and $k \in \{1,2,4,8,16\}$, we sample class-balanced $k$-shot
support sets from the training split and reuse the same indices across all methods.

Training on the target dataset uses the backbone’s standard train-time augmentations
(random resized crops and horizontal flips followed by CLIP-style normalization),
while evaluation uses the standard validation preprocessing (resize, center crop, normalization).

Unless otherwise stated, sigma-only fine-tuning for BOLT uses:
\begin{itemize}
  \item Optimizer: AdamW on all sigma parameters.
  \item Learning rate: $1\times 10^{-3}$.
  \item Weight decay: $0$.
  \item Epochs: $20$.
  \item Batch size: $32$.
  \item Schedule: cosine learning-rate decay with a warmup of 2 epochs.
\end{itemize}

During few-shot training, all encoder weights remain frozen and only the sigma
coefficients in the spectral basis are updated using a standard cross-entropy
loss on the $k$-shot labeled examples. For each target dataset, all remaining datasets from the same domain are used to construct the spectral basis unless otherwise noted.

%-----------------------------------------------------------
\subsection{OOD Training Settings}
%-----------------------------------------------------------
We use the same optimizer, learning rate, weight decay, epoch count, and batch size as in the few-shot setting.
All methods share the same 16-shot support sets and identical training schedules.

%-----------------------------------------------------------
\subsection{Test-Time Adaptation Settings}
%-----------------------------------------------------------

For test-time adaptation (TTA) we use a fully label-free protocol on a held-out target dataset.
The model receives only unlabeled images from the target split.

\paragraph{Trusted sample mining.}
Using the sigma-initialized model with the chosen $\hat{\alpha}$, we run a forward pass
over the entire target split and collect softmax predictions $p(y \mid x)$ for each image.
Let $C$ be the number of classes and $N$ the number of target samples.
We select a fixed number of high-confidence ``trusted'' samples per class:
for each class $c$ we sort examples by $p(y=c \mid x)$ and keep the top
\[
k_{\text{trusted}} = \min\left(\frac{N/C}{10},\, 100\right)
\]
indices for class $c$.
The union of these indices forms the trusted set $\mathcal{D}_{\text{trusted}}$,
and the complement forms the unlabeled set $\mathcal{D}_{\text{unlabeled}}$.
We also store one-hot targets for all samples based on the argmax predictions of this
initial model; these targets are used only for trusted indices.

\paragraph{Two-stream batch construction.}
During TTA training we use a two-stream batch sampler.
Each mini-batch of size $B$ is constructed by sampling
$B/2$ indices from $\mathcal{D}_{\text{unlabeled}}$ and $B/2$ indices from
$\mathcal{D}_{\text{trusted}}$, with independent random permutations for the two streams.

\paragraph{Weak/strong augmentations.}
We use an asymmetric transform to generate weak and strong views:
\begin{itemize}
  \item \textbf{Weak view} $x^{\text{weak}}$: the standard validation preprocessing of the encoder
        (resize, center crop, normalization).
  \item \textbf{Strong view} $x^{\text{strong}}$: a composition of
        \texttt{RandomResizedCrop} to size $224$ with scale $(0.5, 1.0)$ and bicubic
        interpolation, followed by \texttt{RandomHorizontalFlip} with probability $0.5$,
        and finally the last normalization transforms from the validation pipeline.
\end{itemize}

\paragraph{UFM loss with trusted samples.}
Let $\ell_{\text{UFM}}$ denote the loss.
Given logits from the weak and strong views, $\ell_{\text{UFM}}$ proceeds as follows:
\begin{enumerate}
  \item Compute soft predictions from the weak view
        $q(x) = \operatorname{softmax}(\text{logits}^{\text{weak}})$ and apply a simple
        sharpening by scaling and renormalization:
        $\tilde{q}(x) \propto 0.5 \cdot q(x)$.
  \item For trusted indices, overwrite $\tilde{q}(x)$ with the one-hot targets obtained
        from the initial sigma-initialized model (trusted pseudo-labels).
  \item Define a confidence score $w(x) = \max_c \tilde{q}_c(x)$ and a binary mask
        $m(x) = \mathbf{1}[w(x) > \tau]$ with a fixed threshold $\tau = 0.99$.
  \item Compute the per-sample cross-entropy between the strong-view logits and the
        \emph{soft} pseudo-labels, and weight it by $m(x)$:
        \[
        \ell_{\text{UFM}} = \frac{1}{\sum_x m(x)} \sum_x
        m(x)\; \operatorname{CE}\big(\text{logits}^{\text{strong}}(x),\, \tilde{q}(x)\big).
        \]
\end{enumerate}
Thus, TTA optimizes a single UFM-style loss that uses high-confidence pseudo-labels from the weak view,
with trusted examples anchored to fixed one-hot targets and low-confidence examples masked out.

%-----------------------------------------------------------
\subsection{Pseudo-code for BOLT}
%-----------------------------------------------------------

Below we summarize the BOLT pipeline in two stages: 
(i) offline construction of a shared spectral basis and pooled diagonals, and 
(ii) online adaptation to a new task in spectral coordinates.
The mathematical details are given in the main paper; here we focus on implementation flow.

\paragraph{Offline: shared spectral basis and pooled diagonals.}
\begin{enumerate}
  \item For each source task $i$, obtain a fine-tuned model $\Theta_i$ and
        define the task vector $\Delta_i = \Theta_i - \Theta_0$ with respect
        to the pre-trained CLIP weights $\Theta_0$.
  \item For each layer $\ell$, extract the layer-wise update $M_i^{(\ell)}$
        from $\Delta_i$ (reshaped as a matrix).
  \item For each $M_i^{(\ell)}$, compute a thin SVD and keep the top singular
        directions per task and layer.
  \item Stack all singular directions across tasks and apply whitening-based
        orthogonalization to obtain a shared spectral basis
        $U^{(\ell)}_{\text{orth}}, V^{(\ell)}_{\text{orth}}$ for each layer.
  \item Project each task update $M_i^{(\ell)}$ into the shared basis and
        extract the diagonal coefficients $s_i^{(\ell)}$.
  \item Average the layer-wise diagonals across tasks to obtain a pooled
        initializer $s_{\text{pool}}^{(\ell)}$ for each layer.
\end{enumerate}

\paragraph{Online: new task adaptation in spectral coordinates.}
\begin{enumerate}
  \item On a small held-out subset, perform a short sweep over a set of scalar
        scales $\mathcal{A}$ and select $\hat{\alpha}$ that maximizes train
        accuracy.
  \item Initialize the trainable spectral coefficients as
        $s^{(\ell)}_0 = \hat{\alpha} \, s_{\text{pool}}^{(\ell)}$ for all layers.
  \item For a new task, define the objective (few-shot cross-entropy or TTA loss)
        in terms of the parameters $\{s^{(\ell)}\}$ only.
  \item For each training step:
  \begin{enumerate}
    \item Reconstruct the weight update $\Delta(s)$ from the current diagonals
          $\{s^{(\ell)}\}$ and form $\Theta(s) = \Theta_0 + \Delta(s)$.
    \item Run a forward pass on a mini-batch and compute the loss
          (few-shot or TTA).
    \item Backpropagate gradients into $\{s^{(\ell)}\}$ and update them using
          AdamW.
  \end{enumerate}
  \item After a fixed number of epochs, evaluate the final model $\Theta(s)$
        on the target test split.
\end{enumerate}

%%%%%%%%%%%%%%%%%%%%%%%%%%%%%%%%%%%%%%%%%%%%%%%%%%%%%%%%%%%%
\section{Dataset Details}
%%%%%%%%%%%%%%%%%%%%%%%%%%%%%%%%%%%%%%%%%%%%%%%%%%%%%%%%%%%%

%-----------------------------------------------------------
\subsection{General-Domain Benchmarks}
%-----------------------------------------------------------

The general-domain task pool includes the following 17 datasets:
DTD, GTSRB, MNIST, SVHN, STL10, Oxford-IIIT Pet, Flowers102,
CIFAR100, PCAM, CIFAR10, Food101, Fashion-MNIST, RenderedSST2,
EMNIST, FGVCAircraft, CUB200, and Country211.
We use the official train/validation/test splits whenever available, 
or widely adopted splits from the CLIP and prompt-tuning literature.
All methods share the same splits and $k$-shot support sets.
Details for the general-domain datasets are shown in Table~\ref{tab:general_detail}.

\begin{table}[t]
\centering
\caption{\textbf{General-domain datasets.} 
We report the number of classes and fully fine-tuned validation accuracy for three CLIP backbones.}
\label{tab:general_detail}
\resizebox{\linewidth}{!}{
\small
\begin{tabular}{lcccc}
\toprule
Dataset & Classes & ViT-B/32 & ViT-B/16 & ViT-L/14 \\
\midrule
DTD              & 47  & 78.55 & 82.09 & 85.11 \\
GTSRB            & 43  & 99.92 & 99.92 & 99.96 \\
MNIST            & 10  & 99.56 & 99.50 & 99.70 \\
SVHN             & 10  & 96.38 & 96.76 & 97.24 \\
STL10            & 10  & 98.40 & 99.60 & 99.40 \\
Oxford-IIIT Pet  & 37  & 92.39 & 94.84 & 95.92 \\
Flowers102       & 102 & 95.10 & 97.06 & 99.02 \\
CIFAR100         & 100 & 89.52 & 91.08 & 93.68 \\
PCAM             & 2   & 97.36 & 97.86 & 98.04 \\
CIFAR10          & 10  & 97.88 & 98.42 & 99.06 \\
Food101          & 101 & 84.62 & 89.56 & 93.06 \\
Fashion-MNIST    & 10  & 95.52 & 95.28 & 95.66 \\
RenderedSST2     & 2   & 71.39 & 77.31 & 82.51 \\
EMNIST           & 47  & 99.82 & 99.78 & 99.78 \\
FGVCAircraft     & 100 & 40.65 & 47.28 & 68.11 \\
CUB200           & 200 & 73.56 & 77.37 & 86.35 \\
Country211       & 211 & 21.99 & 27.64 & 38.06 \\
\bottomrule
\end{tabular}
}
\end{table}
%-----------------------------------------------------------
\subsection{Remote-Sensing Benchmarks}
%-----------------------------------------------------------

The remote-sensing task pool includes the following 15 datasets:
AID, CLRS, EuroSAT RGB, MLRSNet, NWPU-RESISC45, Optimal-31,
PatternNet, RS C11, RSD46-WHU, RSI-CB128, RSSCN7, SAT-4,
SIRI-WHU, UC Merced, and WHU-RS19.
All remote-sensing datasets are partitioned into training, validation, and test sets using an 8:1:1 ratio.
Details for the remote-sensing datasets are shown in Table~\ref{tab:rs_detail}.

\begin{table}[t]
\centering
\caption{\textbf{Remote-sensing datasets.} 
We report the number of classes and fully fine-tuned validation accuracy for three CLIP backbones.}
\label{tab:rs_detail}
\resizebox{\linewidth}{!}{
\small
\begin{tabular}{lcccc}
\toprule
Dataset & Classes & ViT-B/32 & ViT-B/16 & ViT-L/14 \\
\midrule
AID             & 10 & 98.50 & 99.00 & 99.17 \\
CLRS            & 25 & 89.43 & 90.60 & 91.10 \\
EuroSAT RGB     & 10 & 98.19 & 98.11 & 99.06 \\
MLRSNet         & 30 & 96.28 & 96.39 & 97.12 \\
NWPU-RESISC45   & 45 & 93.97 & 95.44 & 98.03 \\
Optimal-31      & 31 & 95.16 & 95.94 & 96.51 \\
PatternNet      & 38 & 99.72 & 99.77 & 99.81 \\
RS C11          & 11 & 96.76 & 97.57 & 96.82 \\
RSD46-WHU       & 46 & 89.61 & 90.35 & 91.82 \\
RSI-CB128       & 45 & 99.14 & 99.17 & 99.40 \\
RSSCN7          & 7  & 95.71 & 97.68 & 96.79 \\
SAT-4           & 4  & 96.19 & 99.82 & 99.67 \\
SIRI-WHU        & 12 & 97.29 & 98.54 & 98.58 \\
UC Merced       & 21 & 98.81 & 99.52 & 99.60 \\
WHU-RS19        & 19 & 98.51 & 99.50 & 99.58 \\
\bottomrule
\end{tabular}}

\end{table}

%-----------------------------------------------------------
\subsection{Qualitative Examples}
%-----------------------------------------------------------

\begin{figure*}[t]
  \centering
  \begin{minipage}[t]{0.48\linewidth}
    \centering
    % Replace with your own collage of general-domain images
    \includegraphics[width=\linewidth]{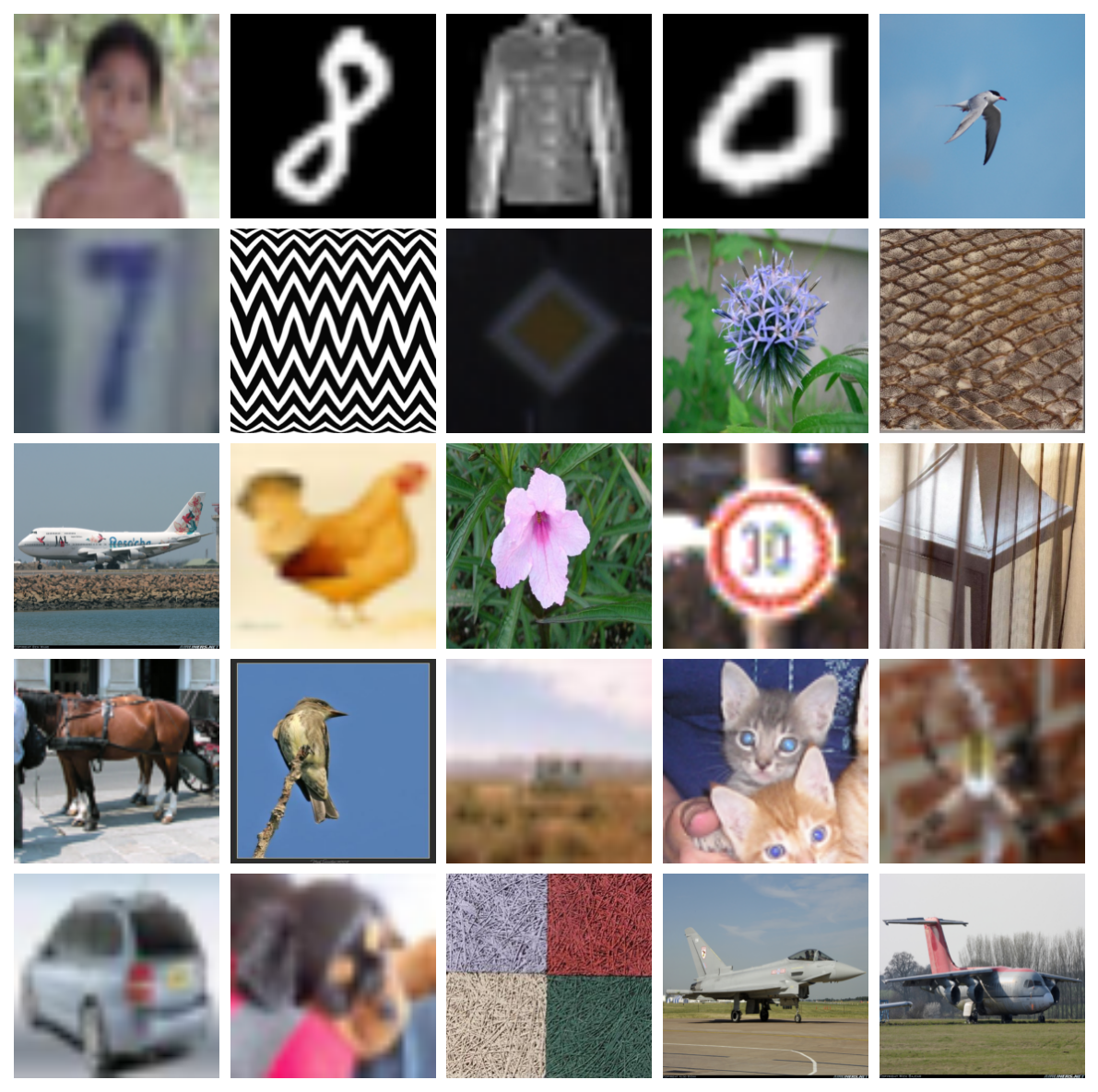}
  \end{minipage}
  \hfill
  \begin{minipage}[t]{0.48\linewidth}
    \centering
    % Replace with your own collage of remote-sensing images
    \includegraphics[width=\linewidth]{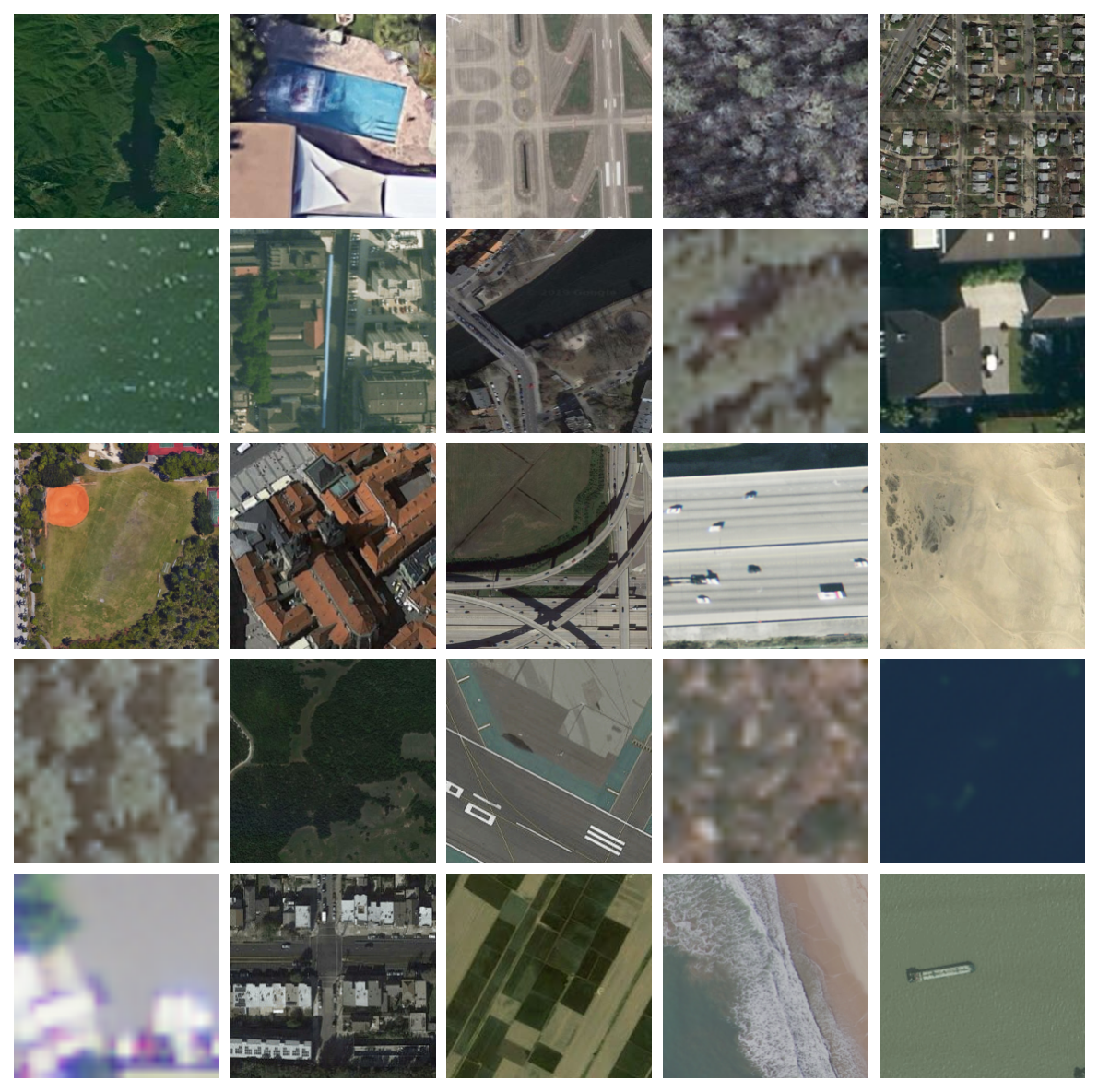}
  \end{minipage}
  \caption{\textbf{Example images from the two domains.}
  General-domain benchmarks (left) and remote-sensing benchmarks (right).}
  \label{fig:dataset_examples}
\end{figure*}

Figure~\ref{fig:dataset_examples} provides qualitative examples from both domains.
The left column shows a collage of general-domain images (textures, digits, traffic signs,
objects, and animals).
The right column shows satellite and aerial scenes from remote-sensing datasets
(urban, agricultural, coastal, and natural regions).

\section{Extended Few-shot Results}

%-----------------------------------------------------------
\subsection{Detailed Few-shot Results on General-Domain Datasets}
%-----------------------------------------------------------

Table~\ref{tab:fewshot_general_detailed} report 
per-dataset few-shot accuracy for all 17 general-domain benchmarks across 
three CLIP backbones (ViT-B/32, ViT-B/16, and ViT-L/14). 
Although a few datasets contain individual settings where certain baselines 
(such as TIP or aTLAS) obtain competitive scores, BOLT achieves the best 
overall performance in the vast majority of cases.  
When averaging across all datasets for each backbone, BOLT consistently 
outperforms all competing methods at every $k \in \{1,2,4,8,16\}$, 
demonstrating strong robustness across domains, dataset sizes, and visual characteristics.

A notable trend is that the performance gap becomes larger as $k$ decreases:
with extremely small support sets (1- or 2-shot), BOLT shows substantial gains over prior parameter-efficient and training-based approaches.  
This confirms that constructing a shared spectral basis and learning only 
task-specific diagonal coefficients is particularly effective in the low-data regime.

%-----------------------------------------------------------
\subsection{Detailed Few-shot Results on Remote-Sensing Datasets}
%-----------------------------------------------------------

Complete few-shot results for all 15 remote-sensing datasets are shown in 
Table~\ref{tab:fewshot_remote_detailed}.  
As in the general-domain setting, BOLT achieves the strongest average accuracy for every backbone and every $k$-shot configuration.  
While a small number of datasets may exhibit close competition with alternative methods, BOLT remains the top-performing approach overall, 
providing stable improvements across diverse Earth-observation tasks.

The advantage of BOLT is again most pronounced in the smallest $k$-shot settings.  
When only one or two labeled samples per class are available, BOLT achieves clear margins over all baselines—highlighting that the proposed 
spectral subspace parameterization captures transferable task structure 
especially well when supervision is extremely limited.

%%%%%%%%%%%%%%%%%%%%%%%%%%%%%%%%%%%%%%%%%%%%%%%%%%%%%%%%%%%%
\section{Extended OOD and TTA Results}
%%%%%%%%%%%%%%%%%%%%%%%%%%%%%%%%%%%%%%%%%%%%%%%%%%%%%%%%%%%%

%-----------------------------------------------------------
\subsection{OOD Results}
%-----------------------------------------------------------
Table \ref{tab:OOD_extend_result} summarizes the additional 16-shot OOD evaluation results on ImageNet variants for ViT-B/16 and ViT-L/14. Across both backbones, BOLT consistently delivers the highest top-1 accuracy on all four OOD datasets—ImageNet-A, ImageNet-R, ImageNet-S, and ImageNet-V2.

For ViT-B/16, BOLT achieves the highest accuracy on all four OOD datasets—ImageNet-A, ImageNet-R, ImageNet-S, and ImageNet-V2—outperforming the next-best method (typically aTLAS) by noticeable margins. In particular, the gains over both linear-probing and adapter-based methods (LoRA, TIP, LP++) are substantial on the more challenging benchmarks such as ImageNet-A and ImageNet-S.

For ViT-L/14, the pattern becomes even stronger. While several baselines achieve competitive second-best numbers on individual datasets, BOLT consistently ranks first across all four OOD datasets, delivering the strongest overall robustness among all evaluated methods. The improvements are especially pronounced on ImageNet-A and ImageNet-R, where the larger backbone benefits significantly from BOLT’s diagonal spectral adaptation.

Overall, these extended results confirm that BOLT’s OOD generalization improvements are not limited to ViT-B/32 but generalize robustly across backbones of different scales. The performance margin tends to enlarge as the model capacity increases, suggesting that BOLT effectively leverages the additional representational power of larger models for robust transfer.

\begin{table*}[t]
\centering
\caption{\textbf{Additional OOD results on ImageNet variants (16-shot setting).}
All entries are top-1 accuracy (\%). Best and second-best per dataset are shown
in \textbf{bold} and \underline{underlined}, respectively.}
\label{tab:OOD_extend_result}
\resizebox{\linewidth}{!}{
\begin{tabular}{lcccccccc}
\toprule
& \multicolumn{4}{c}{ViT-B/16} & \multicolumn{4}{c}{ViT-L/14} \\
\cmidrule(lr){2-5}\cmidrule(lr){6-9}
Method &
ImageNet-A &
ImageNet-R &
ImageNet-S &
ImageNet-V2 &
ImageNet-A &
ImageNet-R &
ImageNet-S &
ImageNet-V2 \\
\midrule
Zero-shot &
\underline{27.67} & 57.53 & 44.22 & 57.88 &
46.36 & 70.30 & 55.38 & 66.52 \\
LinearProbe &
27.44 & 58.31 & 44.91 & 59.83 &
46.47 & 70.69 & 55.78 & 68.02 \\
LoRA &
17.39 & 44.65 & 39.61 & 52.59 &
40.32 & 58.39 & 51.73 & 62.06 \\
TIP &
27.09 & 56.99 & 43.79 & 57.26 &
45.84 & 69.63 & 55.04 & 66.11 \\
LP++ &
22.76 & 50.31 & 41.40 & 50.54 &
\underline{47.43} & 65.88 & 50.15 & 68.97 \\
aTLAS &
26.75 & \underline{59.21} & \underline{45.62} & \underline{60.90} &
47.23 & \underline{73.02} & \underline{57.62} & \underline{69.11} \\
BOLT &
\textbf{28.41} & \textbf{60.91} & \textbf{47.35} & \textbf{61.76} &
\textbf{49.63} & \textbf{74.68} & \textbf{58.69} & \textbf{70.12} \\
\bottomrule
\end{tabular}
}
\end{table*}

%-----------------------------------------------------------
\subsection{Per-dataset TTA Results (General-domain)}
%-----------------------------------------------------------
Table~\ref{tab:tta_general_ufm} reports the full per-dataset test-time adaptation accuracy on all 17 general-domain datasets for the three CLIP backbones ViT-B/32, ViT-B/16, and ViT-L/14. Each entry corresponds to top-1 accuracy after label-free adaptation with the UFM-style objective and provides a more fine-grained analysis beyond the backbone-averaged results.

Across ViT-B/32 and ViT-B/16, BOLT achieves the highest average accuracy and 
delivers strong improvements over Zeroshot CLIP, LayerNorm, and aTLAS baselines. 
This trend is most evident on challenging fine-grained datasets such as 
FGVCAircraft and CUB200, where the structured spectral-diagonal updates appear particularly effective. These results suggest that for smaller CLIP backbones, BOLT is able to capture meaningful task-specific variations even when pseudo-labels are imperfect.

For ViT-L/14, LayerNorm becomes a more competitive baseline—especially on 
difficult datasets—sometimes outperforming BOLT. This behavior is reasonable: 
larger backbones have substantially more parameters, and updating even a small subset via spectral coefficients can be less stable during TTA, whereas 
normalization-only adjustments (as in LayerNorm) remain lightweight and robust. 
Nevertheless, BOLT still achieves the best overall average accuracy among all 
methods on ViT-L/14, indicating that the spectral basis remains effective even at large scale.

Overall, the detailed results in Table~\ref{tab:tta_general_ufm} show that BOLT provides consistent gains across a wide range of datasets and model sizes. The method is especially effective for smaller backbones and for harder tasks, while remaining competitive with stronger normalization-based baselines on larger models. These findings complement the supervised few-shot experiments and demonstrate that BOLT is a robust approach for label-free test-time adaptation without additional supervised fine-tuning.

\begin{table*}[t]
    \centering
    \scriptsize
    \setlength{\tabcolsep}{4.0pt}
    \caption{\textbf{Per-dataset TTA accuracy (\%) on general-domain datasets with ViT-B/32, ViT-B/16, ViT-L/14}
  Each entry reports top-1 accuracy after label-free test-time adaptation using the UFM-style objective.}
    \label{tab:tta_general_ufm}
    \resizebox{\linewidth}{!}{
    \begin{tabular}{l l c c c c c c c c c c c c c c c c c c}
        \toprule
        Backbone & Method & \rot{DTD} & \rot{GTSRB} & \rot{MNIST} & \rot{SVHN} & \rot{STL10} & \rot{OxfordIIITPet} & \rot{Flowers102} & \rot{CIFAR100} & \rot{PCAM} & \rot{CIFAR10} & \rot{Food101} & \rot{FashionMNIST} & \rot{RenderedSST2} & \rot{EMNIST} & \rot{FGVCAircraft} & \rot{CUB200} & \rot{Country211} & \rot{Average} \\
        \midrule
        \multirow{4}{*}{ViT-B-32} & Zeroshot CLIP & 44.41 & 32.56 & 48.25 & 31.61 & 97.12 & 87.44 & 66.32 & 64.20 & \underline{60.62} & 89.83 & 82.72 & 63.02 & 58.59 & 50.24 & 19.56 & 53.00 & 17.21 & 56.87 \\
         & LayerNorm & \underline{46.22} & \underline{41.94} & \underline{60.22} & \underline{59.01} & \underline{97.17} & \underline{89.10} & \underline{66.74} & \textbf{73.29} & \textbf{64.93} & \underline{93.57} & \underline{83.73} & \underline{75.86} & \underline{61.34} & \underline{61.46} & \underline{20.04} & \underline{53.52} & \textbf{17.42} & \underline{62.68} \\
         & aTLAS & 44.41 & 32.56 & 48.25 & 31.61 & 97.12 & 87.44 & 66.32 & 64.20 & \underline{60.62} & 89.83 & 82.72 & 63.02 & 58.59 & 50.24 & 19.56 & 53.00 & 17.21 & 56.87 \\
         & BOLT (ours) & \textbf{48.62} & \textbf{51.21} & \textbf{83.22} & \textbf{75.27} & \textbf{97.67} & \textbf{90.71} & \textbf{69.82} & \underline{72.77} & 58.56 & \textbf{94.13} & \textbf{84.41} & \textbf{80.41} & \textbf{64.20} & \textbf{89.22} & \textbf{20.64} & \textbf{53.57} & \underline{17.24} & \textbf{67.74} \\
        \specialrule{1pt}{2pt}{2pt}
        \multirow{4}{*}{ViT-B/16} & Zeroshot CLIP & 44.68 & 43.34 & 51.79 & 51.98 & 98.25 & 89.04 & 71.15 & 66.91 & 54.02 & 90.80 & 87.68 & 67.32 & 60.52 & 66.43 & 24.30 & 55.37 & 22.84 & 61.55 \\
         & LayerNorm & \underline{45.74} & \underline{49.15} & \underline{75.59} & \underline{70.49} & \underline{98.40} & \underline{91.39} & \underline{71.98} & \underline{75.21} & \textbf{64.37} & \underline{94.40} & \underline{88.39} & \textbf{78.27} & \underline{61.07} & \underline{79.53} & \underline{25.11} & \underline{57.73} & \underline{22.93} & \underline{67.63} \\
         & aTLAS & 44.68 & 43.34 & 51.79 & 51.98 & 98.25 & 89.04 & 71.15 & 66.91 & 54.02 & 90.80 & 87.68 & 67.32 & 60.52 & 66.57 & 24.30 & 55.37 & 22.84 & 61.56 \\
         & BOLT (ours) & \textbf{46.17} & \textbf{50.73} & \textbf{86.83} & \textbf{70.82} & \textbf{98.52} & \textbf{91.52} & \textbf{72.95} & \textbf{75.53} & \underline{63.17} & \textbf{95.43} & \textbf{88.46} & \underline{77.29} & \textbf{62.00} & \textbf{97.84} & \textbf{25.32} & \textbf{57.75} & \textbf{23.12} & \textbf{69.62} \\
        \specialrule{1pt}{2pt}{2pt}
        
        \multirow{4}{*}{ViT-L/14} & Zeroshot CLIP & 55.37 & 50.55 & 76.36 & 58.45 & 99.36 & 93.43 & 79.17 & 75.82 & 51.21 & 95.57 & 92.33 & 66.94 & \underline{68.92} & 65.04 & 31.77 & 62.19 & 31.86 & 67.90 \\
         & LayerNorm & \underline{56.81} & \textbf{58.91} & \textbf{92.19} & \underline{71.29} & \textbf{99.45} & \textbf{94.24} & \underline{79.26} & \textbf{83.96} & \underline{59.11} & \underline{97.52} & \textbf{93.02} & \textbf{79.13} & \textbf{69.36} & \underline{93.95} & \textbf{34.20} & \textbf{64.31} & \textbf{32.60} & \underline{74.08} \\
         & aTLAS & 55.37 & 50.55 & 76.36 & 58.45 & 99.36 & 93.43 & 79.17 & 75.82 & 51.21 & 95.57 & 92.33 & 66.94 & \underline{68.92} & 65.34 & 31.77 & 62.19 & 31.86 & 67.92 \\
         & BOLT (ours) & \textbf{56.86} & \underline{57.74} & \underline{90.50} & \textbf{71.44} & \underline{99.38} & \underline{94.44} & \textbf{79.46} & \underline{83.54} & \textbf{64.01} & \textbf{97.68} & \underline{92.73} & \underline{77.88} & 64.96 & \textbf{98.76} & \underline{34.14} & \underline{64.08} & \underline{32.35} & \textbf{74.11} \\
        \bottomrule
    \end{tabular}
}
\end{table*}

\clearpage
\thispagestyle{empty}
\begin{table*}[t]
\centering
\tiny
\setlength{\tabcolsep}{5.0pt}
\caption{Few-shot general domain accuracy (\%) for each dataset and backbone. Results are reported for $k\in\{1,2,4,8,16\}$.}
\label{tab:fewshot_general_detailed}
\begin{tabular}{l c l c c c c c c c c c c c c c c c c c c}
\toprule
Backbone & $k$ & Method & \rot{DTD} & \rot{GTSRB} & \rot{MNIST} & \rot{SVHN} & \rot{STL10} & \rot{OxfordIIITPet} & \rot{Flowers102} & \rot{CIFAR100} & \rot{PCAM} & \rot{CIFAR10} & \rot{Food101} & \rot{FashionMNIST} & \rot{RenderedSST2} & \rot{EMNIST} & \rot{FGVCAircraft} & \rot{CUB200} & \rot{Country211} & \rot{Average} \\
\midrule

\multirow{30}{*}{ViT-B/32} & \multirow{6}{*}{1} & Linear Probe & 41.97 & 32.87 & 53.95 & 28.79 & \underline{96.21} & 86.94 & 64.32 & 63.00 & \underline{62.80} & 86.80 & \underline{80.55} & 62.87 & 57.72 & 50.40 & 18.39 & 52.57 & 16.36 & 56.26 \\
 &  & LoRA & 43.09 & 41.27 & 60.91 & 41.71 & 93.16 & 85.53 & 47.96 & 63.04 & 56.02 & 83.19 & 76.33 & 61.35 & \underline{59.20} & 24.92 & 11.52 & 32.57 & 13.85 & 52.68 \\
 &  & TIP & 42.93 & 33.05 & 55.98 & 29.14 & 96.17 & \underline{87.19} & \underline{64.69} & 63.25 & 62.70 & 87.08 & 80.53 & 62.78 & 57.06 & 51.92 & 18.72 & \underline{53.04} & \underline{16.41} & 56.63 \\
 &  & LP++ & 31.81 & 32.34 & 42.25 & 13.47 & 68.01 & 51.89 & 63.12 & 31.07 & 59.44 & 60.98 & 41.35 & 52.24 & 53.54 & 39.80 & 13.77 & 21.57 & 3.83 & 40.03 \\
 &  & aTLAS & \underline{45.64} & \underline{48.21} & \underline{85.12} & \underline{63.64} & 95.29 & 83.65 & 63.59 & \underline{70.29} & 62.22 & \underline{89.78} & 78.53 & \underline{70.38} & 58.15 & \underline{93.19} & \underline{19.53} & 51.05 & 13.47 & \underline{64.22} \\
 &  & BOLT (ours) & \textbf{48.24} & \textbf{58.38} & \textbf{86.14} & \textbf{69.70} & \textbf{96.78} & \textbf{88.53} & \textbf{68.86} & \textbf{71.27} & \textbf{68.24} & \textbf{91.50} & \textbf{82.11} & \textbf{73.28} & \textbf{60.41} & \textbf{97.88} & \textbf{21.72} & \textbf{53.52} & \textbf{16.51} & \textbf{67.83} \\
\cmidrule(lr){2-21}
 & \multirow{6}{*}{2} & Linear Probe & 41.97 & 32.88 & 54.04 & 28.79 & 96.21 & 86.92 & 64.32 & 63.00 & 62.81 & 86.83 & 80.55 & 62.87 & \underline{57.61} & 50.40 & 18.42 & 52.69 & 16.37 & 56.28 \\
 &  & LoRA & 45.59 & \underline{53.92} & 66.96 & 40.63 & 96.19 & 85.94 & 55.15 & 67.45 & 63.17 & 87.01 & 77.70 & 62.37 & 53.65 & 17.09 & 8.28 & 43.06 & 13.68 & 55.17 \\
 &  & TIP & 43.72 & 34.35 & 56.90 & 25.75 & 96.14 & \underline{87.24} & 65.62 & 63.42 & 62.46 & 87.28 & 80.53 & 64.22 & 55.96 & 51.04 & 19.29 & 53.09 & \underline{16.54} & 56.68 \\
 &  & LP++ & 45.21 & 46.83 & 69.24 & 18.87 & 86.59 & 63.37 & \textbf{71.02} & 40.75 & \underline{64.58} & 74.34 & 44.45 & 66.68 & 49.37 & 63.83 & 17.73 & 32.97 & 4.78 & 50.62 \\
 &  & aTLAS & \underline{46.38} & 49.96 & \underline{85.68} & \underline{65.07} & \underline{96.54} & 85.20 & 66.40 & \underline{71.70} & 62.31 & \underline{88.66} & \underline{80.78} & \underline{71.12} & 55.41 & \underline{92.23} & \underline{20.94} & \underline{53.23} & 15.59 & \underline{65.13} \\
 &  & BOLT (ours) & \textbf{51.91} & \textbf{64.14} & \textbf{86.23} & \textbf{70.24} & \textbf{97.12} & \textbf{89.32} & \underline{70.45} & \textbf{73.23} & \textbf{68.01} & \textbf{91.18} & \textbf{82.71} & \textbf{74.76} & \textbf{59.80} & \textbf{96.70} & \textbf{24.21} & \textbf{55.06} & \textbf{16.61} & \textbf{68.92} \\
\cmidrule(lr){2-21}
 & \multirow{6}{*}{4} & Linear Probe & 42.02 & 32.88 & 54.00 & 28.80 & 96.21 & 86.92 & 64.37 & 63.03 & 62.81 & 86.83 & 80.56 & 62.87 & \underline{57.61} & 50.45 & 18.36 & 52.68 & 16.38 & 56.28 \\
 &  & LoRA & 48.67 & 59.05 & 74.77 & 41.92 & 95.62 & \underline{88.58} & 66.06 & 67.37 & \underline{74.33} & 85.37 & 79.02 & 64.69 & 55.57 & 22.71 & 4.38 & 48.76 & 14.15 & 58.30 \\
 &  & TIP & 45.32 & 36.14 & 53.90 & 28.65 & \underline{96.25} & 87.46 & 66.38 & 63.88 & 63.28 & 87.57 & 80.66 & 65.88 & 57.28 & 53.27 & 20.04 & 53.37 & \underline{16.46} & 57.40 \\
 &  & LP++ & \underline{50.05} & \underline{59.13} & 77.08 & 17.92 & 91.31 & 61.95 & \textbf{82.21} & 46.56 & 74.10 & 77.79 & 57.21 & 66.36 & 53.10 & 75.05 & \underline{24.03} & 47.13 & 7.20 & 56.95 \\
 &  & aTLAS & 46.49 & 58.08 & \underline{87.89} & \underline{72.50} & 95.58 & 86.89 & 69.10 & \underline{72.89} & 72.52 & \underline{89.06} & \underline{81.43} & \underline{74.31} & 54.09 & \underline{95.70} & 22.68 & \underline{54.12} & 16.41 & \underline{67.63} \\
 &  & BOLT (ours) & \textbf{55.64} & \textbf{71.77} & \textbf{90.35} & \textbf{75.07} & \textbf{97.59} & \textbf{89.59} & \underline{72.53} & \textbf{74.54} & \textbf{75.74} & \textbf{92.30} & \textbf{83.30} & \textbf{77.73} & \textbf{60.74} & \textbf{97.56} & \textbf{26.43} & \textbf{57.04} & \textbf{16.98} & \textbf{71.46} \\
\cmidrule(lr){2-21}
 & \multirow{6}{*}{8} & Linear Probe & 42.07 & 32.94 & 54.08 & 28.86 & 96.20 & 87.00 & 64.42 & 63.10 & 62.80 & 86.86 & 80.59 & 62.91 & 57.61 & 50.50 & 18.45 & 52.73 & 16.36 & 56.32 \\
 &  & LoRA & 55.64 & \underline{71.09} & 79.00 & 54.61 & 96.36 & 88.88 & \underline{82.47} & 72.76 & 66.88 & 89.15 & 80.04 & 38.51 & 56.56 & 17.37 & 6.54 & 55.47 & 14.92 & 60.37 \\
 &  & TIP & 47.23 & 37.59 & 57.28 & 24.78 & 96.30 & 87.41 & 68.22 & 64.50 & 60.65 & 88.49 & \underline{80.61} & 66.46 & 56.23 & 52.57 & 20.16 & 53.81 & 16.55 & 57.58 \\
 &  & LP++ & \underline{58.67} & 68.77 & 85.59 & 25.56 & 95.53 & 71.65 & \textbf{91.53} & 54.79 & 71.14 & 82.14 & 67.73 & 72.73 & 54.64 & 81.83 & \textbf{31.74} & \textbf{60.15} & 9.75 & 63.76 \\
 &  & aTLAS & 51.17 & 65.04 & \underline{88.22} & \underline{72.03} & \underline{96.89} & \underline{88.96} & 71.02 & \underline{74.99} & \underline{73.24} & \textbf{93.06} & \textbf{83.71} & \underline{78.82} & \underline{58.54} & \underline{97.88} & 23.73 & 55.02 & \underline{17.19} & \underline{69.97} \\
 &  & BOLT (ours) & \textbf{60.00} & \textbf{79.90} & \textbf{92.43} & \textbf{79.71} & \textbf{97.46} & \textbf{90.41} & 75.62 & \textbf{76.26} & \textbf{75.52} & \underline{92.89} & \textbf{83.71} & \textbf{80.58} & \textbf{60.52} & \textbf{97.97} & \underline{29.94} & \underline{60.03} & \textbf{17.70} & \textbf{73.57} \\
\cmidrule(lr){2-21}
 & \multirow{6}{*}{16} & Linear Probe & 42.23 & 33.06 & 54.19 & 28.91 & 96.21 & 87.03 & 64.56 & 63.11 & 62.79 & 86.88 & 80.63 & 62.98 & 57.61 & 50.52 & 18.45 & 52.93 & 16.40 & 56.38 \\
 &  & LoRA & 61.76 & \underline{80.46} & 84.37 & 61.01 & 96.16 & 88.61 & \underline{85.09} & \underline{76.30} & 71.60 & 89.83 & 81.75 & 36.00 & \underline{60.13} & 20.87 & 23.22 & \underline{63.77} & 16.35 & 64.55 \\
 &  & TIP & 49.95 & 44.83 & 67.94 & 23.40 & 96.28 & 87.82 & 68.53 & 64.69 & 60.68 & 88.77 & 80.94 & 68.62 & 55.96 & 59.27 & 20.55 & 52.04 & 15.64 & 59.17 \\
 &  & LP++ & \textbf{64.47} & 74.34 & 90.62 & 33.37 & 96.54 & 79.34 & \textbf{91.75} & 61.66 & 71.74 & 85.90 & 74.29 & 76.98 & 54.97 & 87.62 & \textbf{36.45} & \textbf{67.85} & 12.50 & 68.26 \\
 &  & aTLAS & 53.35 & 68.95 & \underline{90.77} & \underline{80.14} & \underline{96.96} & \underline{90.11} & 71.56 & 75.89 & \underline{73.35} & \underline{93.25} & \underline{84.13} & \underline{82.13} & 53.54 & \underline{97.78} & 23.34 & 55.73 & \underline{17.38} & \underline{71.08} \\
 &  & BOLT (ours) & \underline{63.72} & \textbf{82.76} & \textbf{94.87} & \textbf{81.48} & \textbf{97.81} & \textbf{90.65} & 78.76 & \textbf{77.37} & \textbf{76.80} & \textbf{94.07} & \textbf{84.26} & \textbf{83.06} & \textbf{61.34} & \textbf{98.08} & \underline{33.09} & 63.00 & \textbf{18.33} & \textbf{75.26} \\

\specialrule{1pt}{2pt}{2pt}

\multirow{30}{*}{ViT-B/16} & \multirow{6}{*}{1} & Linear Probe & 42.29 & 42.83 & 62.05 & 50.48 & \underline{97.69} & 87.54 & 67.80 & 62.38 & 58.97 & 88.35 & \underline{85.84} & 64.32 & 58.26 & 69.20 & 23.73 & 53.33 & 21.01 & 60.95 \\
 &  & LoRA & \textbf{49.15} & 54.80 & 70.67 & 59.94 & 96.95 & 81.47 & \textbf{78.35} & 68.93 & \underline{63.50} & 90.25 & 84.50 & \underline{72.08} & 54.75 & 81.15 & \underline{26.82} & \textbf{57.78} & 19.30 & 65.32 \\
 &  & TIP & 42.87 & 42.91 & 63.29 & 51.61 & 97.62 & 87.76 & 68.43 & 62.62 & 59.40 & 88.44 & 85.83 & 64.40 & 58.76 & 70.68 & 24.00 & 53.75 & \underline{21.06} & 61.38 \\
 &  & LP++ & 31.70 & 33.64 & 50.08 & 18.42 & 65.94 & 53.12 & 67.38 & 33.06 & 60.69 & 67.44 & 52.18 & 43.52 & 52.11 & 51.17 & 17.88 & 29.88 & 5.13 & 43.14 \\
 &  & aTLAS & 45.11 & \underline{56.46} & \underline{80.08} & \underline{70.81} & 96.66 & \underline{88.61} & 69.82 & \textbf{73.45} & 61.65 & \underline{92.54} & 85.75 & 70.49 & \underline{58.92} & \underline{97.07} & 24.90 & 53.97 & 18.92 & \underline{67.37} \\
 &  & BOLT (ours) & \underline{47.93} & \textbf{62.30} & \textbf{88.13} & \textbf{77.42} & \textbf{98.10} & \textbf{91.44} & \underline{73.62} & \underline{72.89} & \textbf{68.51} & \textbf{92.93} & \textbf{87.65} & \textbf{73.63} & \textbf{61.18} & \textbf{98.92} & \textbf{27.99} & \underline{57.40} & \textbf{21.44} & \textbf{70.67} \\
\cmidrule(lr){2-21}
 & \multirow{6}{*}{2} & Linear Probe & 42.29 & 42.87 & 62.13 & 50.48 & \underline{97.69} & 87.52 & 67.85 & 62.43 & 58.96 & 88.34 & 85.85 & 64.32 & 58.15 & 69.20 & 23.79 & 53.28 & 21.00 & 60.95 \\
 &  & LoRA & \textbf{54.73} & \underline{63.15} & 78.40 & 58.01 & 97.52 & 88.20 & \textbf{84.00} & 71.84 & 60.57 & 90.60 & 86.28 & \textbf{76.36} & \underline{60.68} & 77.16 & \textbf{30.93} & \textbf{60.13} & 19.43 & 68.12 \\
 &  & TIP & 43.94 & 43.40 & 66.61 & 51.86 & 97.64 & 87.79 & 69.12 & 62.63 & 60.16 & 88.64 & 85.89 & 65.13 & 57.83 & 71.56 & 24.60 & 54.09 & \underline{21.25} & 61.89 \\
 &  & LP++ & 45.16 & 49.25 & 69.84 & 22.30 & 87.90 & 71.25 & 75.23 & 43.91 & \underline{62.27} & 75.08 & 53.82 & 62.85 & 51.89 & 70.22 & 23.73 & 40.71 & 6.02 & 53.61 \\
 &  & aTLAS & 47.82 & 61.77 & \underline{79.77} & \underline{70.33} & 97.25 & \underline{89.23} & 71.90 & \textbf{74.62} & 61.98 & \underline{92.82} & \underline{86.47} & 74.29 & 59.53 & \underline{95.25} & 27.00 & 56.02 & 20.61 & \underline{68.63} \\
 &  & BOLT (ours) & \underline{52.87} & \textbf{68.76} & \textbf{87.37} & \textbf{76.87} & \textbf{98.16} & \textbf{91.31} & \underline{77.48} & \underline{74.28} & \textbf{69.07} & \textbf{92.83} & \textbf{88.05} & \underline{75.42} & \textbf{61.72} & \textbf{98.78} & \underline{30.75} & \underline{58.77} & \textbf{21.56} & \textbf{72.00} \\
\cmidrule(lr){2-21}
 & \multirow{6}{*}{4} & Linear Probe & 42.39 & 42.88 & 62.13 & 50.48 & 97.69 & 87.54 & 67.88 & 62.47 & 58.97 & 88.36 & 85.85 & 64.31 & 58.15 & 69.22 & 23.91 & 53.42 & 21.05 & 60.98 \\
 &  & LoRA & \textbf{58.51} & \underline{76.06} & 76.68 & 70.31 & \underline{98.12} & \underline{91.82} & \textbf{88.57} & \underline{74.70} & 73.74 & 90.48 & 86.49 & 74.09 & 56.84 & 80.14 & \textbf{34.14} & \textbf{64.08} & 19.97 & \underline{71.46} \\
 &  & TIP & 44.89 & 43.81 & 70.24 & 51.53 & 97.69 & 87.98 & 70.21 & 62.97 & 61.36 & 88.84 & 86.10 & 66.07 & 58.05 & 72.15 & 25.80 & 55.09 & 21.50 & 62.60 \\
 &  & LP++ & 51.06 & 59.03 & 76.65 & 25.85 & 94.00 & 64.79 & \underline{86.73} & 51.06 & \textbf{75.42} & 78.34 & 64.78 & 63.21 & 57.00 & 82.77 & 30.90 & 55.38 & 8.98 & 60.35 \\
 &  & aTLAS & 48.94 & 63.01 & \textbf{90.48} & \underline{76.68} & 97.75 & 90.38 & 74.22 & 74.43 & 64.99 & \underline{92.20} & \underline{87.66} & \underline{76.88} & \underline{58.65} & \underline{97.77} & 27.66 & 57.92 & \underline{21.56} & 70.66 \\
 &  & BOLT (ours) & \underline{57.34} & \textbf{77.86} & \underline{90.41} & \textbf{80.11} & \textbf{98.32} & \textbf{92.42} & 81.88 & \textbf{75.31} & \underline{75.22} & \textbf{93.51} & \textbf{88.34} & \textbf{78.30} & \textbf{62.38} & \textbf{98.32} & \underline{33.24} & \underline{61.30} & \textbf{22.18} & \textbf{74.50} \\
\cmidrule(lr){2-21}
 & \multirow{6}{*}{8} & Linear Probe & 42.39 & 42.94 & 62.32 & 50.46 & 97.69 & 87.63 & 68.04 & 62.52 & 58.97 & 88.35 & 85.88 & 64.35 & 58.21 & 69.22 & 23.91 & 53.71 & 21.06 & 61.04 \\
 &  & LoRA & \textbf{66.60} & \textbf{83.46} & 82.54 & 69.71 & 97.66 & \underline{92.34} & \textbf{94.44} & \textbf{77.73} & 58.33 & 90.80 & 87.59 & \underline{80.05} & 59.03 & 81.39 & \textbf{42.09} & \textbf{69.73} & 20.25 & \underline{73.75} \\
 &  & TIP & 46.91 & 46.70 & 68.83 & 47.71 & \underline{97.72} & 88.69 & 71.70 & 63.62 & 63.06 & 88.80 & 86.19 & 68.27 & 62.27 & 73.81 & 27.00 & 55.16 & 21.56 & 63.41 \\
 &  & LP++ & 59.10 & 70.00 & 86.83 & 34.41 & 97.25 & 79.34 & \underline{92.37} & 58.10 & \underline{73.27} & 87.46 & 75.48 & 71.35 & 60.08 & 87.30 & \underline{38.31} & \underline{67.24} & 12.28 & 67.66 \\
 &  & aTLAS & 51.60 & 71.16 & \textbf{92.64} & \underline{82.03} & 97.70 & 91.85 & 76.17 & 76.54 & \textbf{74.20} & \underline{94.20} & \underline{88.28} & 79.68 & \textbf{63.21} & \underline{97.93} & 29.88 & 58.35 & \textbf{22.82} & 73.43 \\
 &  & BOLT (ours) & \underline{62.98} & \underline{82.14} & \underline{92.13} & \textbf{83.30} & \textbf{98.50} & \textbf{92.56} & 84.34 & \underline{77.44} & 73.15 & \textbf{94.25} & \textbf{88.74} & \textbf{80.40} & \underline{62.38} & \textbf{98.40} & 36.78 & 64.96 & \underline{22.70} & \textbf{76.19} \\
\cmidrule(lr){2-21}
 & \multirow{6}{*}{16} & Linear Probe & 42.55 & 43.06 & 62.50 & 50.53 & 97.70 & 87.65 & 68.08 & 62.81 & 58.97 & 88.35 & 85.91 & 64.48 & 58.15 & 69.28 & 23.94 & 54.04 & 21.12 & 61.12 \\
 &  & LoRA & \textbf{70.85} & \textbf{88.27} & 87.09 & 79.58 & 97.89 & 92.70 & \textbf{94.97} & \textbf{80.48} & 69.85 & 92.14 & 88.17 & 82.15 & 50.14 & 89.19 & \textbf{48.42} & \textbf{76.03} & 21.36 & \underline{77.02} \\
 &  & TIP & 48.99 & 45.77 & 74.54 & 51.34 & 97.81 & 89.02 & 72.16 & 63.72 & 63.02 & 89.66 & 86.24 & 69.27 & 60.74 & 76.27 & 27.15 & 52.76 & 20.15 & 64.04 \\
 &  & LP++ & 65.64 & 75.47 & 92.45 & 40.23 & 97.84 & 83.86 & \underline{93.33} & 65.46 & \underline{71.08} & 89.82 & 81.07 & 76.49 & 60.68 & 90.70 & \underline{44.94} & \underline{74.25} & 15.38 & 71.69 \\
 &  & aTLAS & 53.72 & 73.98 & \underline{92.73} & \underline{83.06} & \underline{98.17} & \underline{92.86} & 77.18 & 78.04 & 70.68 & \textbf{94.57} & \textbf{89.13} & \underline{83.02} & \underline{62.60} & \underline{98.67} & 22.41 & 59.34 & \underline{22.81} & 73.71 \\
 &  & BOLT (ours) & \underline{66.91} & \underline{85.48} & \textbf{95.64} & \textbf{85.07} & \textbf{98.35} & \textbf{93.05} & 85.72 & \underline{79.04} & \textbf{72.93} & \underline{94.24} & \underline{89.09} & \textbf{83.29} & \textbf{65.68} & \textbf{98.74} & 41.58 & 69.26 & \textbf{23.37} & \textbf{78.09} \\
 
\specialrule{1pt}{2pt}{2pt}
\multirow{30}{*}{ViT-L/14} & \multirow{6}{*}{1} & Linear Probe & 53.30 & 53.97 & 72.12 & 54.43 & \underline{98.90} & 92.67 & 76.26 & 72.77 & 58.25 & 93.48 & 91.03 & 62.28 & 69.19 & 56.43 & 28.62 & 60.46 & 28.43 & 66.03 \\
 &  & LoRA & \textbf{60.48} & \underline{71.23} & \underline{91.94} & \underline{69.64} & 97.44 & 82.47 & \underline{81.33} & 68.81 & 49.49 & 77.72 & 81.26 & 25.35 & 51.29 & \underline{90.25} & 28.20 & 56.25 & 20.29 & 64.91 \\
 &  & TIP & 53.72 & 55.33 & 73.19 & 54.03 & \underline{98.90} & \underline{92.97} & 77.09 & 73.04 & \underline{58.90} & \underline{93.58} & 91.01 & \underline{62.51} & \textbf{70.29} & 56.78 & 29.16 & \underline{61.10} & \underline{28.44} & 66.47 \\
 &  & LP++ & 35.64 & 47.26 & 54.21 & 27.49 & 78.01 & 59.91 & 73.22 & 33.68 & 53.79 & 63.51 & 55.93 & 47.39 & 54.70 & 51.81 & 20.76 & 33.91 & 5.40 & 46.86 \\
 &  & aTLAS & 53.46 & 61.33 & 72.10 & 54.41 & \underline{98.90} & \underline{92.97} & 78.74 & \textbf{81.10} & 58.27 & 93.48 & \underline{91.22} & 62.27 & \underline{69.25} & 56.36 & \underline{34.98} & 60.34 & 27.49 & \underline{67.45} \\
 &  & BOLT (ours) & \underline{56.86} & \textbf{74.11} & \textbf{93.79} & \textbf{80.79} & \textbf{99.14} & \textbf{93.79} & \textbf{83.57} & \underline{80.98} & \textbf{61.23} & \textbf{96.09} & \textbf{91.54} & \textbf{76.26} & 64.31 & \textbf{98.90} & \textbf{37.92} & \textbf{63.53} & \textbf{28.98} & \textbf{75.40} \\
\cmidrule(lr){2-21}
 & \multirow{6}{*}{2} & Linear Probe & 53.30 & 53.97 & 72.13 & \underline{54.43} & \underline{98.90} & 92.67 & 76.27 & 72.80 & 58.28 & 93.49 & \underline{91.03} & \underline{62.29} & 69.19 & 56.41 & 28.59 & 60.55 & 28.44 & 66.04 \\
 &  & LoRA & \textbf{62.66} & \textbf{80.33} & \underline{87.33} & 24.86 & 95.04 & 46.66 & \underline{87.33} & 73.40 & \textbf{72.78} & 90.77 & 82.78 & 16.47 & 55.35 & 49.83 & 27.60 & 62.46 & 19.73 & 60.91 \\
 &  & TIP & 54.04 & 53.97 & 73.88 & 53.17 & \underline{98.90} & 92.91 & 77.80 & 73.18 & 59.82 & \underline{93.65} & \underline{91.03} & 61.98 & \textbf{71.11} & 56.95 & 30.60 & 61.62 & \underline{28.52} & 66.66 \\
 &  & LP++ & 47.13 & 62.40 & 75.57 & 31.38 & 92.30 & 77.27 & \textbf{88.01} & 49.67 & 51.32 & 79.58 & 64.55 & 60.25 & 52.17 & \underline{68.69} & 29.61 & 49.86 & 7.64 & 58.08 \\
 &  & aTLAS & 56.70 & 67.21 & 72.10 & 54.41 & \underline{98.90} & \underline{93.65} & 79.66 & \underline{81.54} & 58.27 & 93.48 & 90.88 & 62.27 & \underline{69.25} & 56.36 & \underline{36.18} & \underline{62.65} & 28.29 & \underline{68.34} \\
 &  & BOLT (ours) & \underline{59.89} & \underline{77.24} & \textbf{91.36} & \textbf{80.80} & \textbf{99.11} & \textbf{94.30} & 86.78 & \textbf{81.69} & \underline{70.63} & \textbf{95.70} & \textbf{91.98} & \textbf{76.56} & 67.49 & \textbf{98.55} & \textbf{41.40} & \textbf{67.38} & \textbf{29.34} & \textbf{77.07} \\
\cmidrule(lr){2-21}
 & \multirow{6}{*}{4} & Linear Probe & 53.30 & 54.08 & 72.18 & 54.43 & 98.90 & 92.67 & 76.27 & 72.85 & 58.31 & 93.50 & 91.02 & 62.33 & 69.19 & 56.46 & 28.71 & 60.58 & 28.43 & 66.07 \\
 &  & LoRA & \textbf{66.49} & \textbf{84.87} & 88.21 & 38.16 & 95.78 & 91.09 & \textbf{92.36} & 76.45 & \underline{66.30} & 80.16 & 64.61 & 71.86 & 57.61 & 25.64 & 9.87 & \textbf{72.13} & 20.56 & 64.83 \\
 &  & TIP & 55.48 & 56.66 & 73.93 & 53.96 & \underline{98.92} & 93.00 & 78.81 & 73.42 & 60.49 & 93.65 & 91.13 & 63.51 & \textbf{71.22} & 57.43 & 31.83 & 62.20 & 28.81 & 67.32 \\
 &  & LP++ & 52.39 & 71.28 & 81.59 & 37.81 & 96.30 & 75.99 & 91.38 & 59.57 & 63.85 & 88.08 & 75.78 & 67.20 & 57.33 & 80.12 & 37.17 & 63.84 & 11.15 & 65.34 \\
 &  & aTLAS & 59.20 & 75.42 & \underline{89.90} & \underline{78.10} & 98.41 & \underline{93.02} & 84.36 & \textbf{82.96} & 58.27 & \underline{95.41} & \underline{91.68} & \underline{77.03} & \underline{69.25} & \textbf{98.98} & \underline{37.98} & 66.00 & \underline{29.88} & \underline{75.64} \\
 &  & BOLT (ours) & \underline{64.15} & \underline{82.95} & \textbf{93.08} & \textbf{81.57} & \textbf{99.16} & \textbf{94.52} & \underline{91.53} & \underline{82.78} & \textbf{76.07} & \textbf{96.00} & \textbf{92.17} & \textbf{77.76} & 66.67 & \underline{98.19} & \textbf{41.55} & \underline{70.37} & \textbf{30.19} & \textbf{78.75} \\
\cmidrule(lr){2-21}
 & \multirow{6}{*}{8} & Linear Probe & 53.40 & 54.16 & 72.20 & 54.46 & 98.90 & 92.72 & 76.40 & 72.95 & 58.29 & 93.49 & 91.04 & 62.36 & 69.19 & 56.51 & 28.89 & 60.70 & 28.47 & 66.13 \\
 &  & LoRA & \textbf{69.89} & \textbf{86.29} & 11.11 & 34.75 & 96.67 & 86.73 & \underline{96.18} & 80.19 & 57.50 & 90.26 & 87.10 & 73.09 & 63.65 & 17.82 & 2.07 & \textbf{79.06} & 19.98 & 61.90 \\
 &  & TIP & 56.01 & 59.39 & 76.67 & 52.48 & \underline{98.95} & 92.97 & 80.05 & 73.60 & 63.67 & 93.72 & 91.30 & 64.55 & \underline{70.07} & 61.81 & 32.34 & 62.56 & 28.67 & 68.17 \\
 &  & LP++ & 63.78 & 81.21 & 89.33 & 43.46 & 97.61 & 84.49 & \textbf{96.29} & 69.18 & \underline{68.44} & 90.07 & 82.69 & 70.17 & 63.15 & 86.86 & \textbf{45.87} & \underline{76.27} & 16.21 & 72.06 \\
 &  & aTLAS & 60.48 & 78.56 & \underline{92.01} & \underline{79.16} & 98.67 & \underline{93.43} & 88.47 & \underline{83.76} & 58.27 & \underline{96.45} & \underline{92.32} & \underline{78.61} & 69.25 & \underline{98.39} & 39.66 & 67.24 & \textbf{31.58} & \underline{76.84} \\
 &  & BOLT (ours) & \underline{65.11} & \underline{86.16} & \textbf{94.34} & \textbf{84.93} & \textbf{99.28} & \textbf{94.58} & 94.41 & \textbf{84.04} & \textbf{75.12} & \textbf{96.55} & \textbf{92.55} & \textbf{81.36} & \textbf{70.13} & \textbf{98.76} & \underline{44.73} & 74.63 & \underline{30.42} & \textbf{80.42} \\
\cmidrule(lr){2-21}
 & \multirow{6}{*}{16} & Linear Probe & 53.46 & 54.28 & 72.22 & 54.50 & 98.89 & 92.72 & 76.42 & 73.11 & 58.33 & 93.50 & 91.07 & 62.45 & \underline{69.19} & 56.58 & 28.95 & 60.98 & 28.51 & 66.19 \\
 &  & LoRA & \textbf{75.53} & \textbf{90.70} & 20.88 & 10.43 & 98.14 & 91.47 & \textbf{97.19} & 81.72 & 70.07 & 90.69 & 87.85 & 79.61 & \textbf{69.25} & 16.55 & \textbf{56.53} & \textbf{82.91} & 22.07 & 67.15 \\
 &  & TIP & 57.34 & 63.03 & 79.13 & 53.33 & \underline{98.99} & \underline{93.32} & 80.37 & 73.92 & 66.22 & 93.89 & 91.41 & 66.78 & 64.85 & 68.67 & 18.96 & 59.98 & 27.97 & 68.13 \\
 &  & LP++ & 70.37 & 85.61 & 93.34 & 50.40 & 98.90 & 89.42 & \underline{96.49} & 74.12 & \underline{71.61} & 94.79 & 87.80 & 76.84 & 60.90 & 90.66 & \underline{53.98} & \underline{81.31} & 20.89 & 76.32 \\
 &  & aTLAS & 63.83 & 82.54 & \underline{95.02} & \textbf{86.51} & 98.79 & \textbf{94.96} & 88.18 & \textbf{85.42} & 58.27 & \underline{96.63} & \underline{92.87} & \underline{82.51} & \textbf{69.25} & \underline{98.52} & 41.19 & 68.55 & \textbf{32.06} & \underline{78.53} \\
 &  & BOLT (ours) & \underline{71.06} & \underline{88.27} & \textbf{96.30} & \underline{86.05} & \textbf{99.25} & \textbf{94.96} & 94.96 & \underline{85.15} & \textbf{76.15} & \textbf{97.24} & \textbf{92.97} & \textbf{83.87} & \underline{69.19} & \textbf{98.87} & 49.53 & 79.27 & \underline{31.27} & \textbf{82.02} \\
\bottomrule
\end{tabular}
\end{table*}

\clearpage
\thispagestyle{empty}
\begin{table*}[t]
\centering
\tiny
\setlength{\tabcolsep}{5.0pt}
\caption{Few-shot remote-sensing accuracy (\%) for each dataset and backbone. Results are reported for $k\in\{1,2,4,8,16\}$.}
\label{tab:fewshot_remote_detailed}
\begin{tabular}{l c l c c c c c c c c c c c c c c c c}
\toprule
Backbone & $k$ & Method & \rot{AID} & \rot{CLRS} & \rot{EuroSAT\_RGB} & \rot{MLRSNet} & \rot{NWPU-RESISC45} & \rot{Optimal-31} & \rot{PatternNet} & \rot{RSD46-WHU} & \rot{RSI-CB128} & \rot{RSSCN7} & \rot{RS\_C11} & \rot{SAT-4} & \rot{SIRI-WHU} & \rot{UC\_Merced} & \rot{WHU-RS19} & \rot{Average} \\
\midrule

\multirow{30}{*}{ViT-B/32} & \multirow{6}{*}{1} & Linear Probe & 71.67 & 55.33 & 50.37 & 55.71 & 58.59 & 69.62 & 60.90 & 29.34 & 28.74 & 54.11 & 51.42 & 44.95 & 45.83 & 62.38 & 83.58 & 54.84 \\
 &  & LoRA & \underline{81.50} & 62.97 & \underline{60.06} & 65.51 & 66.27 & 78.76 & \underline{79.41} & 26.57 & 44.42 & 57.68 & \underline{78.14} & 70.45 & \underline{67.29} & 76.90 & 86.57 & 66.83 \\
 &  & TIP & 72.00 & 55.70 & 53.65 & 57.10 & 59.57 & 70.43 & 63.60 & 31.14 & 29.60 & 54.11 & 53.44 & 46.87 & 47.50 & 64.05 & 84.58 & 56.22 \\
 &  & LP++ & 66.67 & 33.40 & 53.11 & 44.12 & 41.62 & 54.30 & 62.86 & 27.14 & \underline{51.78} & 43.21 & 57.09 & 41.16 & 51.46 & 51.67 & 61.69 & 49.42 \\
 &  & aTLAS & 77.83 & \underline{71.77} & 58.65 & \underline{72.85} & \underline{78.95} & \underline{92.20} & 78.31 & \underline{39.67} & 44.91 & \underline{61.07} & 74.49 & \underline{74.02} & \underline{67.29} & \underline{87.86} & \underline{92.54} & \underline{71.49} \\
 &  & BOLT (ours) & \textbf{92.83} & \textbf{76.73} & \textbf{79.30} & \textbf{78.68} & \textbf{81.79} & \textbf{92.74} & \textbf{81.97} & \textbf{46.83} & \textbf{54.96} & \textbf{77.50} & \textbf{78.95} & \textbf{83.89} & \textbf{76.46} & \textbf{89.76} & \textbf{96.52} & \textbf{79.26} \\
\cmidrule(lr){2-19}
 & \multirow{6}{*}{2} & Linear Probe & 71.67 & 55.43 & 50.37 & 55.83 & 58.67 & 69.62 & 61.28 & 29.48 & 28.82 & 54.11 & 51.82 & 44.89 & 45.83 & 62.38 & 83.58 & 54.92 \\
 &  & LoRA & 84.83 & 67.50 & \underline{69.93} & 70.30 & 69.86 & 81.72 & \underline{85.87} & 38.81 & 59.48 & 62.50 & \underline{82.19} & 65.00 & 65.62 & 85.95 & 89.05 & 71.91 \\
 &  & TIP & 74.00 & 56.73 & 55.54 & 58.01 & 60.76 & 71.24 & 65.64 & 34.50 & 31.24 & 54.29 & 56.28 & 48.20 & 49.79 & 66.43 & 86.07 & 57.91 \\
 &  & LP++ & 79.33 & 51.07 & 69.65 & 57.53 & 62.51 & 69.35 & 79.52 & \underline{46.26} & \textbf{69.50} & 59.82 & 71.26 & 55.22 & 60.42 & 68.81 & 79.60 & 65.32 \\
 &  & aTLAS & \underline{88.67} & \underline{73.47} & 63.87 & \underline{75.95} & \underline{80.06} & \textbf{94.62} & 78.42 & 41.27 & 47.04 & \underline{64.82} & 76.52 & \underline{67.56} & \underline{70.62} & \underline{90.00} & \underline{98.01} & \underline{74.06} \\
 &  & BOLT (ours) & \textbf{96.00} & \textbf{78.97} & \textbf{84.24} & \textbf{79.13} & \textbf{82.14} & \underline{93.82} & \textbf{87.40} & \textbf{49.91} & \underline{60.90} & \textbf{80.00} & \textbf{87.85} & \textbf{85.67} & \textbf{82.08} & \textbf{93.57} & \textbf{98.51} & \textbf{82.68} \\
\cmidrule(lr){2-19}
 & \multirow{6}{*}{4} & Linear Probe & 71.67 & 55.50 & 50.48 & 56.04 & 59.05 & 70.16 & 61.45 & 30.31 & 29.13 & 54.11 & 52.23 & 45.20 & 45.83 & 62.38 & 83.58 & 55.14 \\
 &  & LoRA & \underline{92.17} & 71.30 & \underline{77.87} & 77.40 & 78.33 & 86.56 & \textbf{90.56} & 51.83 & 59.23 & 73.75 & 84.21 & \underline{83.48} & \underline{80.21} & 89.52 & 96.02 & \underline{79.50} \\
 &  & TIP & 75.33 & 58.07 & 58.63 & 60.62 & 63.05 & 73.92 & 69.31 & 37.56 & 34.62 & 54.64 & 62.35 & 53.99 & 53.54 & 69.29 & 88.06 & 60.86 \\
 &  & LP++ & 87.50 & 61.40 & 69.43 & 64.75 & 64.83 & 73.12 & 83.75 & \underline{53.60} & \textbf{78.17} & \underline{75.89} & \underline{86.23} & 79.66 & 73.75 & 81.19 & 94.03 & 75.15 \\
 &  & aTLAS & 91.83 & \underline{76.60} & 74.46 & \underline{78.94} & \underline{83.24} & \textbf{95.70} & 81.58 & 45.80 & 49.10 & 63.04 & 84.21 & 79.59 & 77.08 & \underline{91.19} & \textbf{98.51} & 78.06 \\
 &  & BOLT (ours) & \textbf{96.33} & \textbf{81.67} & \textbf{89.30} & \textbf{81.83} & \textbf{84.63} & \underline{94.89} & \underline{90.12} & \textbf{54.20} & \underline{70.10} & \textbf{85.18} & \textbf{91.50} & \textbf{89.42} & \textbf{87.92} & \textbf{93.57} & \underline{98.01} & \textbf{85.91} \\
\cmidrule(lr){2-19}
 & \multirow{6}{*}{8} & Linear Probe & 71.67 & 56.07 & 50.94 & 56.49 & 59.75 & 70.70 & 62.19 & 30.99 & 29.68 & 54.29 & 53.04 & 45.14 & 45.83 & 62.86 & 84.58 & 55.61 \\
 &  & LoRA & 95.17 & 75.77 & \underline{86.30} & \underline{82.18} & 81.62 & 90.32 & \textbf{94.56} & 57.71 & \underline{84.19} & 80.36 & \underline{90.69} & \underline{89.45} & \underline{85.21} & \underline{94.05} & 95.02 & \underline{85.51} \\
 &  & TIP & 78.67 & 61.37 & 63.41 & 64.85 & 65.94 & 77.42 & 75.33 & 42.87 & 38.57 & 52.50 & 70.45 & 56.19 & 58.13 & 69.76 & 90.55 & 64.40 \\
 &  & LP++ & \underline{95.67} & 71.80 & 78.80 & 74.97 & 75.25 & 80.38 & 90.95 & \textbf{63.90} & \textbf{86.56} & \underline{82.14} & 90.28 & 83.41 & 81.04 & 83.33 & 93.53 & 82.13 \\
 &  & aTLAS & 92.67 & \underline{78.87} & 77.98 & 81.09 & \underline{84.48} & \underline{96.77} & 84.44 & 46.78 & 54.52 & 70.18 & 85.43 & 76.57 & 83.54 & 92.86 & \underline{98.01} & 80.28 \\
 &  & BOLT (ours) & \textbf{97.17} & \textbf{82.20} & \textbf{92.07} & \textbf{84.12} & \textbf{85.71} & \textbf{97.31} & \underline{93.34} & \underline{58.45} & 80.97 & \textbf{90.00} & \textbf{93.12} & \textbf{91.00} & \textbf{91.25} & \textbf{95.71} & \textbf{98.51} & \textbf{88.73} \\
\cmidrule(lr){2-19}
 & \multirow{6}{*}{16} & Linear Probe & 72.83 & 56.70 & 51.02 & 57.64 & 60.92 & 71.51 & 63.31 & 32.91 & 30.63 & 54.64 & 55.47 & 46.44 & 46.88 & 64.76 & 84.58 & 56.68 \\
 &  & LoRA & \underline{97.17} & 78.93 & \underline{92.00} & \underline{85.92} & \underline{85.70} & 92.47 & \textbf{96.61} & \textbf{71.58} & \textbf{93.64} & \underline{86.43} & \underline{94.33} & \underline{92.51} & \underline{91.67} & \underline{96.19} & \underline{98.51} & \underline{90.24} \\
 &  & TIP & 83.17 & 64.53 & 67.22 & 70.10 & 69.02 & 80.65 & 81.20 & 47.12 & 46.49 & 62.32 & 72.87 & 60.34 & 68.54 & 75.24 & 92.04 & 69.39 \\
 &  & LP++ & 96.17 & 75.33 & 83.46 & 81.98 & 80.14 & 87.90 & \underline{95.07} & \underline{70.89} & \underline{91.17} & 85.54 & 92.31 & 88.24 & 89.38 & 92.86 & 97.01 & 87.16 \\
 &  & aTLAS & 96.17 & \underline{79.43} & 88.17 & 82.70 & 84.78 & \underline{96.51} & 85.87 & 48.49 & 56.13 & 85.36 & 91.50 & 88.72 & 86.46 & 94.76 & 98.01 & 84.20 \\
 &  & BOLT (ours) & \textbf{97.83} & \textbf{83.97} & \textbf{93.52} & \textbf{86.76} & \textbf{87.60} & \textbf{97.04} & 94.77 & 62.41 & 85.03 & \textbf{91.96} & \textbf{94.74} & \textbf{93.08} & \textbf{94.17} & \textbf{96.90} & \textbf{99.50} & \textbf{90.62} \\

\specialrule{1pt}{2pt}{2pt}

\multirow{30}{*}{ViT-B/16} & \multirow{6}{*}{1} & Linear Probe & 74.83 & 60.17 & 50.02 & 62.77 & 64.16 & 71.77 & 63.50 & 33.30 & 28.26 & 50.36 & 63.56 & 46.81 & 53.54 & 62.86 & 84.08 & 58.00 \\
 &  & LoRA & \underline{88.33} & \underline{66.63} & \underline{74.26} & 72.75 & 71.95 & 76.61 & \underline{84.46} & 34.53 & 50.35 & \underline{67.50} & \underline{82.19} & \underline{65.75} & 47.08 & 79.29 & 89.55 & 70.08 \\
 &  & TIP & 75.50 & 60.83 & 53.41 & 63.81 & 65.19 & 73.39 & 65.84 & 35.36 & 30.36 & 50.89 & 65.18 & 51.90 & 55.42 & 63.81 & 86.07 & 59.80 \\
 &  & LP++ & 69.00 & 44.47 & 56.85 & 51.56 & 52.95 & 61.02 & 67.04 & 29.54 & \underline{55.20} & 30.00 & 64.37 & 45.57 & 50.00 & 60.24 & 63.18 & 53.40 \\
 &  & aTLAS & 82.83 & 51.03 & 59.83 & \underline{77.55} & \underline{78.19} & \textbf{92.47} & 78.29 & \underline{44.15} & 46.27 & 65.00 & 79.76 & 63.22 & \underline{68.54} & \underline{85.00} & \underline{96.52} & \underline{71.24} \\
 &  & BOLT (ours) & \textbf{91.50} & \textbf{77.40} & \textbf{79.20} & \textbf{79.18} & \textbf{83.25} & \underline{91.40} & \textbf{86.50} & \textbf{51.86} & \textbf{56.97} & \textbf{73.04} & \textbf{91.09} & \textbf{77.86} & \textbf{76.88} & \textbf{89.29} & \textbf{97.01} & \textbf{80.16} \\
\cmidrule(lr){2-19}
 & \multirow{6}{*}{2} & Linear Probe & 74.83 & 60.17 & 50.04 & 62.79 & 64.21 & 71.77 & 63.67 & 33.45 & 28.28 & 50.36 & 63.56 & 46.47 & 53.54 & 62.86 & 85.07 & 58.07 \\
 &  & LoRA & \underline{89.17} & 56.33 & \underline{82.83} & 74.48 & 76.65 & 83.60 & \underline{90.12} & \underline{49.26} & 58.23 & 59.82 & 49.39 & 57.89 & \underline{72.71} & 83.57 & 71.14 & 70.35 \\
 &  & TIP & 76.33 & 61.30 & 58.26 & 64.66 & 66.32 & 75.81 & 68.32 & 38.01 & 32.69 & 52.86 & 64.78 & 54.19 & 56.25 & 64.29 & 87.06 & 61.41 \\
 &  & LP++ & 70.50 & 58.10 & 69.98 & 61.27 & 66.97 & 71.77 & 81.07 & 47.20 & \textbf{69.19} & 61.25 & \underline{80.16} & 55.89 & 62.92 & 68.10 & 85.57 & 67.33 \\
 &  & aTLAS & 80.33 & \underline{74.90} & 51.69 & \underline{77.89} & \underline{80.73} & \textbf{94.89} & 81.17 & 45.55 & 50.64 & \underline{73.39} & 76.52 & \underline{64.13} & 70.62 & \underline{88.81} & \underline{96.52} & \underline{73.85} \\
 &  & BOLT (ours) & \textbf{93.50} & \textbf{80.37} & \textbf{85.50} & \textbf{80.27} & \textbf{85.06} & \underline{92.74} & \textbf{90.38} & \textbf{54.59} & \underline{64.90} & \textbf{82.14} & \textbf{89.47} & \textbf{73.20} & \textbf{84.38} & \textbf{93.81} & \textbf{98.01} & \textbf{83.22} \\
\cmidrule(lr){2-19}
 & \multirow{6}{*}{4} & Linear Probe & 75.00 & 60.40 & 50.17 & 62.96 & 64.59 & 72.58 & 64.14 & 33.82 & 28.49 & 50.36 & 63.97 & 46.38 & 53.54 & 62.86 & 85.07 & 58.29 \\
 &  & LoRA & \underline{94.50} & \underline{73.83} & \underline{88.46} & 77.44 & 73.98 & 87.37 & \underline{91.35} & \underline{57.28} & \underline{75.93} & \underline{75.54} & 56.28 & 86.22 & 68.33 & 86.19 & 97.01 & \underline{79.31} \\
 &  & TIP & 78.83 & 63.00 & 62.00 & 66.18 & 68.16 & 76.61 & 72.19 & 41.24 & 38.15 & 56.07 & 68.83 & 58.55 & 58.33 & 67.14 & 89.05 & 64.29 \\
 &  & LP++ & 93.17 & 62.50 & 75.59 & 69.27 & 69.25 & 76.08 & 84.80 & 56.51 & \textbf{75.97} & 71.61 & 86.23 & \underline{86.50} & \underline{79.38} & 76.67 & 88.06 & 76.77 \\
 &  & aTLAS & 84.50 & 60.13 & 63.48 & \underline{79.93} & \underline{83.06} & \underline{94.62} & 83.98 & 47.92 & 55.54 & 74.29 & \underline{87.04} & 45.12 & 75.62 & \underline{90.24} & \underline{98.51} & 74.93 \\
 &  & BOLT (ours) & \textbf{96.17} & \textbf{82.87} & \textbf{90.33} & \textbf{81.67} & \textbf{87.22} & \textbf{95.16} & \textbf{92.94} & \textbf{61.07} & 74.72 & \textbf{87.32} & \textbf{94.74} & \textbf{88.33} & \textbf{88.12} & \textbf{96.90} & \textbf{99.50} & \textbf{87.80} \\
\cmidrule(lr){2-19}
 & \multirow{6}{*}{8} & Linear Probe & 75.00 & 60.73 & 50.63 & 63.40 & 65.06 & 73.66 & 65.38 & 35.05 & 29.42 & 50.18 & 63.97 & 46.27 & 53.75 & 63.10 & 85.57 & 58.74 \\
 &  & LoRA & \underline{95.00} & \underline{79.77} & \textbf{91.69} & \textbf{84.60} & 80.75 & 90.05 & \textbf{95.89} & \textbf{67.35} & \underline{80.92} & \underline{86.79} & \textbf{95.14} & \underline{84.78} & \underline{88.96} & 89.05 & \underline{98.51} & \underline{87.28} \\
 &  & TIP & 82.50 & 64.57 & 66.30 & 69.44 & 71.10 & 79.57 & 77.65 & 45.83 & 45.36 & 59.46 & 72.06 & 63.20 & 61.46 & 71.90 & 90.55 & 68.06 \\
 &  & LP++ & 94.67 & 72.00 & 80.41 & 78.35 & 79.44 & 83.33 & 92.58 & \underline{65.33} & \textbf{86.64} & 84.82 & 92.31 & 84.61 & 82.71 & 85.71 & 89.05 & 83.46 \\
 &  & aTLAS & 87.50 & 70.63 & 83.11 & 81.25 & \underline{81.63} & \underline{96.51} & 86.94 & 50.91 & 61.10 & 78.39 & 71.26 & 83.06 & 82.92 & \underline{94.76} & 97.01 & 80.47 \\
 &  & BOLT (ours) & \textbf{96.83} & \textbf{83.90} & \underline{91.67} & \underline{84.27} & \textbf{89.21} & \textbf{97.04} & \underline{95.23} & 64.36 & 80.32 & \textbf{89.64} & \underline{94.33} & \textbf{91.52} & \textbf{92.08} & \textbf{97.86} & \textbf{99.00} & \textbf{89.82} \\
\cmidrule(lr){2-19}
 & \multirow{6}{*}{16} & Linear Probe & 75.33 & 61.33 & 51.19 & 64.05 & 66.27 & 74.19 & 66.58 & 36.90 & 31.35 & 50.71 & 65.18 & 46.62 & 54.58 & 63.81 & 86.57 & 59.64 \\
 &  & LoRA & 95.33 & \underline{82.70} & \textbf{94.24} & \textbf{87.70} & 86.06 & 93.82 & \textbf{97.86} & \textbf{76.03} & \textbf{93.57} & \underline{90.71} & \underline{95.14} & 88.31 & \underline{93.96} & 96.43 & \textbf{99.50} & \underline{91.42} \\
 &  & TIP & 87.17 & 69.63 & 70.54 & 72.81 & 75.10 & 82.80 & 83.77 & 49.94 & 54.75 & 69.11 & 80.57 & 70.31 & 65.42 & 78.57 & 93.03 & 73.57 \\
 &  & LP++ & \underline{96.50} & 77.60 & 85.57 & 84.14 & 85.38 & 88.98 & 96.00 & \underline{71.89} & \underline{92.25} & 84.29 & 89.47 & \underline{89.46} & 90.62 & 91.90 & 96.52 & 88.04 \\
 &  & aTLAS & 95.33 & 82.10 & 87.52 & 83.46 & \underline{86.22} & \textbf{97.58} & 90.25 & 52.43 & 68.14 & 85.54 & 91.09 & 89.44 & 87.50 & \underline{96.67} & \underline{99.00} & 86.15 \\
 &  & BOLT (ours) & \textbf{97.50} & \textbf{85.13} & \underline{93.26} & \underline{87.03} & \textbf{90.14} & \underline{97.31} & \underline{96.61} & 70.49 & 86.24 & \textbf{91.43} & \textbf{97.57} & \textbf{94.35} & \textbf{94.58} & \textbf{98.10} & \textbf{99.50} & \textbf{91.95} \\

\specialrule{1pt}{2pt}{2pt}
\multirow{30}{*}{ViT-L/14} & \multirow{6}{*}{1} & Linear Probe & 74.33 & 66.67 & 50.00 & 67.49 & 69.16 & 75.27 & 74.92 & 39.21 & 39.01 & 58.04 & 63.97 & 63.27 & 58.33 & 72.38 & 84.58 & 63.77 \\
 &  & LoRA & \underline{85.00} & \underline{67.67} & \underline{69.22} & \underline{78.00} & 74.54 & \underline{85.75} & 81.35 & \underline{51.11} & \underline{59.02} & \underline{68.21} & \underline{75.71} & \underline{66.22} & \underline{66.04} & \underline{80.24} & \underline{93.53} & \underline{73.44} \\
 &  & TIP & 75.67 & 67.43 & 52.39 & 68.35 & 69.83 & 75.54 & 76.41 & 41.24 & 40.30 & 60.18 & 65.18 & 64.01 & 59.58 & 73.81 & 86.07 & 65.07 \\
 &  & LP++ & 66.00 & 43.33 & 54.46 & 47.97 & 50.37 & 43.55 & 70.41 & 39.55 & 48.68 & 33.57 & 58.30 & 63.00 & 51.25 & 55.95 & 64.68 & 52.74 \\
 &  & aTLAS & 74.33 & 66.57 & 49.69 & 67.18 & \underline{83.95} & 74.73 & \underline{84.01} & 50.63 & 52.23 & 57.86 & 63.56 & 63.09 & 58.13 & 72.38 & 84.58 & 66.86 \\
 &  & BOLT (ours) & \textbf{91.00} & \textbf{81.90} & \textbf{79.91} & \textbf{83.39} & \textbf{88.56} & \textbf{96.24} & \textbf{87.99} & \textbf{57.73} & \textbf{59.83} & \textbf{73.57} & \textbf{90.69} & \textbf{77.85} & \textbf{80.00} & \textbf{95.95} & \textbf{98.51} & \textbf{82.88} \\
\cmidrule(lr){2-19}
 & \multirow{6}{*}{2} & Linear Probe & 74.33 & 66.63 & 50.20 & 67.62 & 69.24 & 75.54 & 74.97 & 39.64 & 39.09 & 58.04 & 63.56 & 63.19 & 58.33 & 72.62 & 85.07 & 63.87 \\
 &  & LoRA & \underline{90.83} & 68.97 & \underline{83.56} & 76.25 & 81.25 & 86.29 & \underline{89.34} & \textbf{63.81} & \textbf{77.51} & \underline{73.93} & \underline{83.40} & 63.54 & \underline{72.92} & 86.19 & 94.53 & \underline{79.49} \\
 &  & TIP & 76.83 & 68.50 & 57.19 & 69.36 & 70.81 & 76.08 & 78.17 & 43.04 & 41.58 & 63.57 & 67.21 & \underline{64.89} & 59.58 & 74.76 & 89.05 & 66.71 \\
 &  & LP++ & 83.50 & 66.03 & 65.74 & 71.48 & 69.75 & 69.62 & 86.88 & 54.48 & \underline{68.78} & 68.21 & 78.95 & 53.94 & 64.79 & 65.00 & 84.58 & 70.11 \\
 &  & aTLAS & 74.33 & \underline{81.33} & 49.69 & \underline{82.22} & \underline{85.19} & \textbf{97.31} & 84.62 & 52.00 & 55.19 & 57.86 & 63.56 & 63.09 & 58.13 & \underline{94.29} & \underline{98.01} & 73.12 \\
 &  & BOLT (ours) & \textbf{93.33} & \textbf{84.37} & \textbf{89.33} & \textbf{84.06} & \textbf{88.98} & \underline{96.51} & \textbf{91.28} & \underline{61.90} & 66.41 & \textbf{83.21} & \textbf{92.71} & \textbf{72.97} & \textbf{83.12} & \textbf{97.62} & \textbf{99.50} & \textbf{85.69} \\
\cmidrule(lr){2-19}
 & \multirow{6}{*}{4} & Linear Probe & 74.33 & 66.60 & 50.41 & 67.83 & 69.57 & 76.08 & 75.38 & 40.27 & 39.35 & 58.04 & 64.37 & 63.32 & 58.96 & 72.62 & 85.07 & 64.15 \\
 &  & LoRA & \underline{86.67} & 65.30 & 76.61 & 82.58 & 85.48 & \underline{90.32} & \underline{90.84} & \underline{63.24} & \textbf{86.90} & \textbf{89.64} & 91.09 & \underline{81.59} & \underline{87.71} & 94.76 & 96.52 & \underline{84.62} \\
 &  & TIP & 80.00 & 69.80 & 62.89 & 71.23 & 72.60 & 77.69 & 79.75 & 46.23 & 44.18 & 66.61 & 71.66 & 67.46 & 63.54 & 77.62 & 93.53 & 69.65 \\
 &  & LP++ & \textbf{95.83} & 70.10 & \underline{80.56} & 74.11 & 75.21 & 83.60 & 88.98 & 58.28 & \underline{77.30} & 81.79 & 89.88 & 79.25 & 76.46 & 82.62 & 88.56 & 80.17 \\
 &  & aTLAS & 84.50 & \underline{81.50} & 79.67 & \underline{83.99} & \underline{87.65} & \textbf{97.85} & 88.72 & 57.62 & 59.15 & 57.86 & \underline{91.50} & 63.09 & 82.71 & \underline{95.00} & \underline{98.01} & 80.59 \\
 &  & BOLT (ours) & \textbf{95.83} & \textbf{85.20} & \textbf{89.30} & \textbf{84.16} & \textbf{90.37} & \textbf{97.85} & \textbf{94.11} & \textbf{66.15} & 74.56 & \underline{87.32} & \textbf{95.55} & \textbf{89.06} & \textbf{89.17} & \textbf{97.38} & \textbf{100.00} & \textbf{89.07} \\
\cmidrule(lr){2-19}
 & \multirow{6}{*}{8} & Linear Probe & 74.67 & 67.27 & 50.76 & 68.47 & 70.24 & 76.88 & 76.86 & 41.52 & 39.99 & 58.21 & 64.78 & 63.42 & 58.75 & 73.10 & 85.07 & 64.67 \\
 &  & LoRA & \underline{88.50} & 81.40 & \underline{88.41} & 82.23 & 87.94 & 91.40 & \underline{95.67} & \textbf{78.54} & \textbf{92.14} & 83.39 & \textbf{97.17} & 71.03 & \underline{82.92} & 75.24 & 96.02 & 86.13 \\
 &  & TIP & 82.67 & 71.67 & 67.89 & 74.34 & 75.67 & 80.65 & 82.96 & 49.69 & 48.23 & 70.00 & 76.11 & 69.86 & 65.83 & 82.38 & \underline{97.01} & 73.00 \\
 &  & LP++ & \textbf{97.83} & 78.13 & 86.06 & 82.38 & 83.59 & 84.68 & 93.63 & 68.81 & \underline{86.73} & \underline{85.18} & 92.71 & \underline{87.77} & 82.08 & 90.24 & 96.02 & \underline{86.39} \\
 &  & aTLAS & 83.00 & \underline{82.77} & 77.83 & \underline{85.48} & \underline{89.25} & \textbf{99.46} & 91.91 & 61.59 & 66.49 & 77.86 & 93.12 & 63.09 & 82.71 & \underline{96.90} & \textbf{99.50} & 83.40 \\
 &  & BOLT (ours) & \textbf{97.83} & \textbf{85.23} & \textbf{93.11} & \textbf{86.54} & \textbf{91.63} & \underline{98.12} & \textbf{95.86} & \underline{71.89} & 80.73 & \textbf{89.11} & \underline{96.76} & \textbf{90.64} & \textbf{92.71} & \textbf{98.57} & \textbf{99.50} & \textbf{91.22} \\
\cmidrule(lr){2-19}
 & \multirow{6}{*}{16} & Linear Probe & 75.17 & 68.07 & 51.67 & 69.41 & 71.67 & 77.69 & 78.08 & 43.64 & 41.02 & 58.93 & 66.80 & 63.69 & 60.42 & 73.33 & 87.06 & 65.78 \\
 &  & LoRA & \underline{98.17} & 80.90 & \underline{92.65} & \textbf{89.48} & \underline{91.16} & 94.89 & \textbf{97.62} & \textbf{81.88} & \textbf{95.75} & \underline{88.57} & 83.00 & 84.46 & 86.04 & 96.19 & 98.51 & \underline{90.62} \\
 &  & TIP & 85.67 & 74.27 & 73.02 & 77.96 & 78.75 & 84.14 & 86.23 & 54.02 & 55.75 & 75.54 & 79.35 & 75.24 & 69.58 & 84.76 & 98.01 & 76.82 \\
 &  & LP++ & \textbf{98.33} & 82.77 & 90.41 & 86.60 & 87.02 & 90.05 & 96.04 & \underline{77.05} & \underline{92.44} & 86.61 & 92.71 & \underline{90.30} & \underline{90.62} & 94.29 & 98.01 & 90.22 \\
 &  & aTLAS & 91.33 & \underline{85.27} & 88.59 & 86.64 & 90.33 & \underline{98.66} & 93.34 & 64.47 & 71.45 & 85.89 & \underline{96.76} & 86.09 & 89.38 & \underline{97.86} & \underline{99.50} & 88.37 \\
 &  & BOLT (ours) & \textbf{98.33} & \textbf{86.90} & \textbf{95.17} & \underline{88.60} & \textbf{92.70} & \textbf{99.19} & \underline{96.45} & 75.80 & 85.67 & \textbf{91.25} & \textbf{97.98} & \textbf{93.16} & \textbf{94.38} & \textbf{99.52} & \textbf{100.00} & \textbf{93.01} \\
\bottomrule
\end{tabular}
\end{table*}


\end{document}